\documentclass[letterpaper,twocolumn,10pt]{article}
\usepackage{usenix-2020-09}
\usepackage{microtype}
\microtypecontext{spacing=nonfrench}

\usepackage{listings}
\usepackage[varqu,scale=1.10]{inconsolata}
\definecolor{keywordcolor}{RGB}{0,0,180}
\definecolor{sortcolor}{RGB}{120,0,120}
\definecolor{tacticcolor}{RGB}{0,100,0}
\definecolor{commentcolor}{RGB}{100,100,100}
\definecolor{symbolcolor}{RGB}{180,0,0}
\definecolor{lemmanamecolor}{RGB}{0,128,128}

\usepackage{etoolbox}
\AtBeginEnvironment{thebibliography}{%
  \renewcommand{\ttfamily}{\fontfamily{cmtt}\selectfont}%
}
\lstdefinelanguage{isabelle}{
  morecomment=[l][\color{commentcolor}]{--},
  morecomment=[s][\color{commentcolor}]{(*}{*)},
  morestring=[b]",
  showstringspaces=false,
  keepspaces=true,
  morekeywords=[1]{
    theory, imports, begin, end, context, locale, interpretation,
    lemma, theorem, corollary, proposition, definition, fun, function,
    primrec, datatype, record, abbreviation, type_synonym,
    inductive, coinductive, inductive_set,
    class, instance, instantiation, primcorec, code_datatype,
    proof, qed, show, have, assume, fix, obtain, note, moreover, ultimately,
    using, by, unfolding, sorry, done, oops,
    let, if, then, else, case, of, where, assumes, defines,
    syntax, translations, notation, axiomatization,
    section, subsection, text, txt, declare, schematic_goal,
    interpretation, datatype_compat, value, ML, method_setup
  },
  morekeywords=[2]{bool, nat, int, real, set, list, option, unit, prod, sum},
  morekeywords=[3]{
    simp, auto, blast, force, rule, erule, drule, frule,
    cases, induct, clarify, fastforce, meson, metis,
    subst, unfold, fold, intro, elim, dest, simp_all,
    transfer, presburger, arith, linarith,
    smt, try, nitpick, quickcheck, wp
  },
  morekeywords=[4]{valid_objs, tcb_cap_valid, cap_swap_def, cap_swap, hoare_seq_ext, set_cap_valid_objs},
  literate=
    {α}{{\ensuremath{\alpha}}}1
    {β}{{\ensuremath{\beta}}}1
    {γ}{{\ensuremath{\gamma}}}1
    {δ}{{\ensuremath{\delta}}}1
    {ε}{{\ensuremath{\varepsilon}}}1
    {θ}{{\ensuremath{\theta}}}1
    {λ}{{\ensuremath{\lambda}}}1
    {μ}{{\ensuremath{\mu}}}1
    {ν}{{\ensuremath{\nu}}}1
    {π}{{\ensuremath{\pi}}}1
    {σ}{{\ensuremath{\sigma}}}1
    {τ}{{\ensuremath{\tau}}}1
    {φ}{{\ensuremath{\varphi}}}1
    {ψ}{{\ensuremath{\psi}}}1
    {ω}{{\ensuremath{\omega}}}1
    {∧}{{\ensuremath{\wedge}}}1
    {∨}{{\ensuremath{\vee}}}1
    {¬}{{\ensuremath{\neg}}}1
    {⇒}{{\ensuremath{\Rightarrow}}}1
    {→}{{\ensuremath{\rightarrow}}}1
    {←}{{\ensuremath{\leftarrow}}}1
    {↔}{{\ensuremath{\leftrightarrow}}}1
    {∀}{{\ensuremath{\forall}}}1
    {∃}{{\ensuremath{\exists}}}1
    {≠}{{\ensuremath{\neq}}}1
    {≤}{{\ensuremath{\le}}}1
    {≥}{{\ensuremath{\ge}}}1
    {⊆}{{\ensuremath{\subseteq}}}1
    {⊂}{{\ensuremath{\subset}}}1
    {∈}{{\ensuremath{\in}}}1
    {∉}{{\ensuremath{\notin}}}1
    {⊢}{{\ensuremath{\vdash}}}1
    {⟹}{{\ensuremath{\Longrightarrow}}}1
    {⟷}{{\ensuremath{\Longleftrightarrow}}}1
    {∅}{{\ensuremath{\emptyset}}}1
    {⟨}{{\ensuremath{\langle}}}1
    {⟩}{{\ensuremath{\rangle}}}1
    {⟦}{{\ensuremath{\llbracket}}}1
    {⟧}{{\ensuremath{\rrbracket}}}1
    {⦃}{{\ensuremath{\lbrace\!|}}}1
    {⦄}{{\ensuremath{|\!\rbrace}}}1
    {↦}{{\ensuremath{\mapsto}}}1
    {↝}{{\ensuremath{\leadsto}}}1
    {∷}{{\ensuremath{::}}}1
    {⋅}{{\ensuremath{\cdot}}}1
    {·}{{\ensuremath{\cdot}}}1
    {…}{{\ensuremath{\ldots}}}1
    {∘}{{\ensuremath{\circ}}}1
    {⊓}{{\ensuremath{\sqcap}}}1
    {⊔}{{\ensuremath{\sqcup}}}1
    {⋃}{{\ensuremath{\bigcup}}}1
    {⋂}{{\ensuremath{\bigcap}}}1
  ,
  basicstyle=\ttfamily\footnotesize\linespread{1.08}\selectfont,
  identifierstyle={\ttfamily\color{black}},
  keywordstyle=[1]{\ttfamily\color{keywordcolor}},
  keywordstyle=[2]{\ttfamily\color{sortcolor}},
  keywordstyle=[3]{\ttfamily\color{tacticcolor}},
  keywordstyle=[4]{\ttfamily\color{lemmanamecolor}},
  commentstyle={\ttfamily\footnotesize\color{commentcolor}},
  stringstyle=\ttfamily,
  tabsize=3,
  breaklines=true,
  breakatwhitespace=true,
  columns=flexible,
  basewidth=0.52em,
  captionpos=b,
  extendedchars=false,
  sensitive=true
}

\lstdefinestyle{isa}{
  language=isabelle,
  frame=single,
  breakindent=0pt,
  aboveskip=0.6em,
  belowskip=0.5em,
}
\lstdefinestyle{isainline}{
  language=isabelle,
  frame=single,
  breakindent=0pt,
  basicstyle=\ttfamily,
}

\usepackage[T1]{fontenc}
\usepackage{verbatim}
\usepackage{algorithm}
\usepackage{algpseudocode}

\usepackage{amsthm}
\usepackage{threeparttable}
\usepackage{amsmath}
\usepackage{amsfonts}
\usepackage{algorithm}
\usepackage{algpseudocode}
\algtext*{EndIf}
\algtext*{EndFor}
\usepackage{xspace}
\usepackage{makecell}
\usepackage{enumitem}
\usepackage{subcaption}
\usepackage{natbib}
\usepackage{booktabs}
\usepackage{multirow}
\usepackage[table]{xcolor}
\usepackage{pifont}
\usepackage{enumitem}
\usepackage{tabularx}

\usepackage[many]{tcolorbox}    	
\tcbuselibrary{listings,skins,breakable}

\definecolor{codebg}{HTML}{F5F5F5}
\definecolor{codeframe}{HTML}{CCCCCC}
\definecolor{cprimary}{HTML}{2166AC}
\definecolor{csecondary}{HTML}{B2182B}
\definecolor{ctertiary}{HTML}{2CA02C}
\newcommand{\isabasicstyle}{\ttfamily\fontsize{7.8}{9.5}\selectfont}
\tcbset{
  isabase/.style={
    listing only,
    listing options={style=isa, frame=none, numbers=none,
      basicstyle=\isabasicstyle,
      columns=fullflexible, keepspaces=true,
      backgroundcolor=\color{white},
      aboveskip=0pt, belowskip=0pt},
    colback=white,
    boxrule=0pt,
    leftrule=2pt,
    arc=0pt,
    left=4pt, right=2pt, top=2pt, bottom=2pt,
  },
}
\newtcblisting{isacode}{
  isabase,
  colframe=cprimary,
  before skip=0.6em,
  after skip=0.4em,
}
\newtcblisting{isacode-blue}{
  isabase,
  colframe=cprimary,
  colback=cprimary!5,
  listing options={style=isa, frame=none, numbers=none,
    basicstyle=\isabasicstyle,
    columns=fullflexible, keepspaces=true,
    backgroundcolor=\color{cprimary!5},
    aboveskip=0pt, belowskip=0pt},
  before skip=0.6em,
  after skip=0.5em,
}
\newtcblisting{isacode-green}{
  isabase,
  colframe=ctertiary,
  colback=ctertiary!5,
  listing options={style=isa, frame=none, numbers=none,
    basicstyle=\isabasicstyle,
    columns=fullflexible, keepspaces=true,
    backgroundcolor=\color{ctertiary!5},
    aboveskip=0pt, belowskip=0pt},
  before skip=0.6em,
  after skip=0.5em,
}
\newtcblisting{isacode-red}{
  isabase,
  colframe=csecondary,
  colback=csecondary!5,
  listing options={style=isa, frame=none, numbers=none,
    basicstyle=\isabasicstyle,
    columns=fullflexible, keepspaces=true,
    backgroundcolor=\color{csecondary!5},
    aboveskip=0pt, belowskip=0pt},
  before skip=0.6em,
  after skip=0.5em,
}
\newtcblisting{isacode-gray}{
  isabase,
  colframe=codeframe,
  colback=codebg,
  listing options={style=isa, frame=none, numbers=none,
    basicstyle=\isabasicstyle,
    columns=fullflexible, keepspaces=true,
    backgroundcolor=\color{codebg},
    aboveskip=0pt, belowskip=0pt},
  before skip=0.6em,
  after skip=0.5em,
}
\usepackage{multicol}  

\definecolor{deepgreen}{RGB}{0,100,0}   

\usepackage{tikz}
\usepackage{amsmath}

\usepackage{filecontents}
\usepackage{balance}

\newcommand{\hide}[1]{}

\newtcolorbox{boxA}{
    fontupper = \bf,
    boxrule = 1.5pt,
    left=1mm,
    colframe = black 
}

\begin{document}

\date{}

\newif\ifsubmission
\submissionfalse

\title{Neuro-Symbolic Proof Generation for Scaling Systems Software Verification}

\author{
    {\rm Baoding He}$^{1,2,*}$ \quad 
    {\rm Zenan Li}$^{3,*}$ \quad 
    {\rm Wei Sun}$^{1,2}$ \quad 
    {\rm Yuan Yao}$^{1,2, \dagger}$ \quad 
    {\rm Taolue Chen}$^{4}$ \\[0.5em]
    {\rm Xiaoxing Ma}$^{1,2,\dagger}$ \quad 
    {\rm Zhendong Su}$^{3}$
    \\[0.8em]
    $^{1}${\rm State Key Laboratory of Novel Software Technology, Nanjing University, China}\\
    $^{2}${\rm School of Computer Science, Nanjing University, China} \\
    $^{3}${\rm Department of Computer Science, ETH Zurich, Switzerland} \\
    $^{4}${\rm School of Computing and Mathematical Sciences, Birkbeck, University of London, UK}
}
\renewcommand{\thefootnote}{\fnsymbol{footnote}}
\maketitle
\footnotetext[1]{These authors contributed equally to this work.}
\footnotetext[2]{Corresponding authors.}

\begin{abstract}

Formal verification via interactive theorem proving is increasingly used to ensure the correctness of critical systems, yet constructing large proof scripts remains highly manual and limits scalability. Advances in large language models (LLMs), especially in mathematical reasoning, make their integration into software verification increasingly promising. This paper introduces a neuro-symbolic proof generation framework designed to automate proof search for system-level verification projects. The framework performs a best-first tree search over proof states, repeatedly querying an LLM for the next candidate proof step. On the neural side, we fine-tune LLMs using datasets of proof state–step pairs; on the symbolic side, we incorporate a range of ITP tools to repair rejected steps, filter and rank proof states, and automatically discharge subgoals when search progress stalls. This synergy enables data-efficient LLM adaptation and semantics-informed pruning of the search space. We implement the framework on a new Isabelle REPL that exposes fine-grained proof states and automation tools, and evaluate it on the FVEL seL4 benchmark and additional Isabelle developments. On seL4, the system proves up to 77.6\% of the theorems, substantially surpassing previous LLM-based approaches and standalone Sledgehammer, while solving significantly more multi-step proofs. Results across further Isabelle benchmarks demonstrate strong generalization, indicating a viable path toward scalable automated software verification.
\end{abstract}


\section{Introduction}

Formal verification plays a vital role in ensuring the correctness and reliability of software systems, especially in safety- and security-critical domains where failures can be catastrophic~\citep{Clarke1996, Blanchet2003, Woodcock2009}. Interactive theorem proving (ITP) stands out for its expressiveness and ability to provide the strongest assurance. 
With ITPs, landmark projects such as CompCert (a formally verified optimizing C compiler whose correctness is mechanically proven in Rocq~\cite{leroy2016compcert}) and seL4 (a formally verified OS microkernel whose functional correctness, security properties and reliability guarantees are proven in Isabelle/HOL~\cite{klein2014comprehensive}) have been successfully delivered, which demonstrate that formally verified systems can achieve both practical usability and rigorous correctness.

While ITP offers unparalleled assurance, it also incurs substantial costs, preventing its wide adoption, especially in large-scale, system-level software~\citep{ringer2019qed}. 
Specifically, a typical software verification workflow
involves 1) writing precise formal {\em specifications} to be proved and 2) constructing rigorous {\em proofs} for these specifications, 
both of which demand immense human effort. 
For example, the seL4 verification project required approximately 20 person-years of work, and produced over 100K lines of proof scripts, compared to only about 10K lines of C implementation and merely 3K lines in its simplest abstract specification in Isabelle~\cite{klein2014comprehensive}. 
Moreover, developing such specifications and proofs necessitates specialized expertise in both theorem proving and the target domain, a combination that is challenging to acquire. 

In this work, we focus on the proof generation step in system-level software verification projects. Recently, 
large language models (LLMs) have demonstrated encouraging mathematical reasoning capabilities, which inspire 
research into their application for software verification, particularly for proof generation ~\cite{qin2025can,thompson2024rango,zhang2024selene,lin2024fvel}. 
Unfortunately, both prior studies and our initial experiments show that, even when theorem statements are well formalized and relevant libraries are fully provided in advance, current LLMs still struggle to generate complete, correct proofs. We summarize two challenges as follows.

\noindent{\em \ding{182} Lack of specialized expertise}.
The first challenge lies in that software verification often relies on numerous domain-specific lemmas and specialized proof tactics.
Although modern LLMs perform well in general reasoning tasks, they are not well-versed in these specialized domains, leading to limited effectiveness when applied to proving corresponding theorems.
For example, LLM4FSCQ~\cite{qin2025can} evaluated models such as GPT-4o and Gemini 1.5 Flash on FSCQ (a Rocq-based file system verification project~\cite{fscq}) and achieved only 38\% proof coverage. 
Moreover, in the larger seL4 project, which involves extensive lemma libraries and customized tactics, Selene~\cite{zhang2024selene} found that GPT-4 succeeded on only 20\% of a benchmark theorem set, despite the fact that nearly 40\% of them require one-line proofs.

\noindent{\em \ding{183} Paucity of usable data}.
The second challenge, which is closely related to the first, is the shortage of high-quality (training) data for software verification. 
For instance, the seL4 project contains only around 20K theorems and lemmas, and most proofs are written in a procedural style, where much of the reasoning is implicit and encoded in the interaction with the theorem prover. This substantially limits their usefulness in effective LLM training/fine-tuning.
Among existing seL4 benchmarks, Selene~\cite{zhang2024selene}  extracts a small subset of the project, targeting relatively easy theorems with proof lengths of one to five.
FVEL~\cite{lin2024fvel} is more comprehensive and challenging, as it extracts all theorems to be proved from seL4. Yet even with systematic fine-tuning of Mistral-7B-Instruct~\cite{jiang2023mistral} and Llama-3-8B-Instruct~\cite{dubey2024llama}, FVEL reported a success rate below 10\% (88 out of 1,077 theorems in the test set).

\smallskip
\noindent\textbf{Approach.}
In this paper, we propose a proof generation framework that synergistically combines LLMs and ITPs, aiming for automated system-level software verification. Our approach adopts a proof step-based tree-search pipeline that iteratively queries a fine-tuned LLM to predict the next proof step until the target theorem is successfully proved. 

To address the first challenge,
our framework adopts a neuro-symbolic approach tailored to the nature of software verification. 
First, to address challenges posed by domain-specific lemmas and tactics, we incorporate a proof-step revision module that repairs or refines LLM-generated steps whenever they fail to compile or make no progress. 
Second, to efficiently prune unpromising branches, we utilize Nitpick~\citep{blanchette2010nitpick} and QuickCheck~\citep{claessen2000quickcheck} to “test’’ each candidate proof state and discard those that admit counterexamples; the remaining candidates are then ranked by the LLM according to their plausibility. 
Finally, if the tree search still fails to resolve the goal, Sledgehammer~\citep{bohme2010sledgehammer} acts as a backstop that scans the local library for potentially relevant lemmas and attempts to complete the current proof.


To address the second challenge, we construct our training data by extracting internal proof states from existing theorems.
For example, a five-step proof yields five distinct training instances, which are subsequently used for fine-tuning.
During inference, our tree-search pipeline leverages this state extraction to facilitate direct interaction between the LLM and the prover, informing step predictions.
While we currently train on existing human-authored proofs, this pipeline also enables future ``bootstrapping'', where new training data could be extracted on-the-fly from successful search paths~\citep{wu2024internlm2, xin2025bfs}.

\smallskip

\noindent\textbf{Experiments.} We evaluate our framework on seL4~\cite{klein2014comprehensive} proof construction. The results show that, with a fine-tuned language model, our approach achieves a total proof success rate of up to 77.6\%, significantly surpassing prior automated proof-generation approaches for seL4. Notably, it excels, compared to the two preceding systems, at successfully handling a larger number of multi-step proofs, a challenge that has not been addressed by prior work. 
Moreover, our framework substantially alleviates manual proof-engineering effort, reducing expert proof-writing effort by 71.1\% on average in an AI–human collaboration setting.
Finally, experimental results on benchmarks from Archive of Formal Proofs projects (e.g., X86 Semantics
, IEEE Floating Point, and SATSolverVerification), as well as the code verification benchmark Code2Inv (by translating its loop invariants into Isabelle theorems) confirm the strong generalizability of our approach. 

\smallskip
\noindent\textbf{Summary of main contributions.} 
(1) We provide a neuro-symbolic framework based on LLM-enabled tree search, tailored for real-world, system-level software verification projects. 
(2) We implement concrete integration of LLMs and 
a broad range of symbolic, logic-based tools, substantially enhancing the efficacy of proof-tree search; (3) We curate training datasets and benchmarks, and carry out thorough empirical evaluations. Our work advances a new frontier of AI4Verification, an emerging area focused on applying cutting-edge AI techniques to hardware/software verification, especially to support large-scale proof engineering for safety-critical systems and architectures. 

\begin{figure*}[thb]
    \centering
    \includegraphics[width=0.98\textwidth]{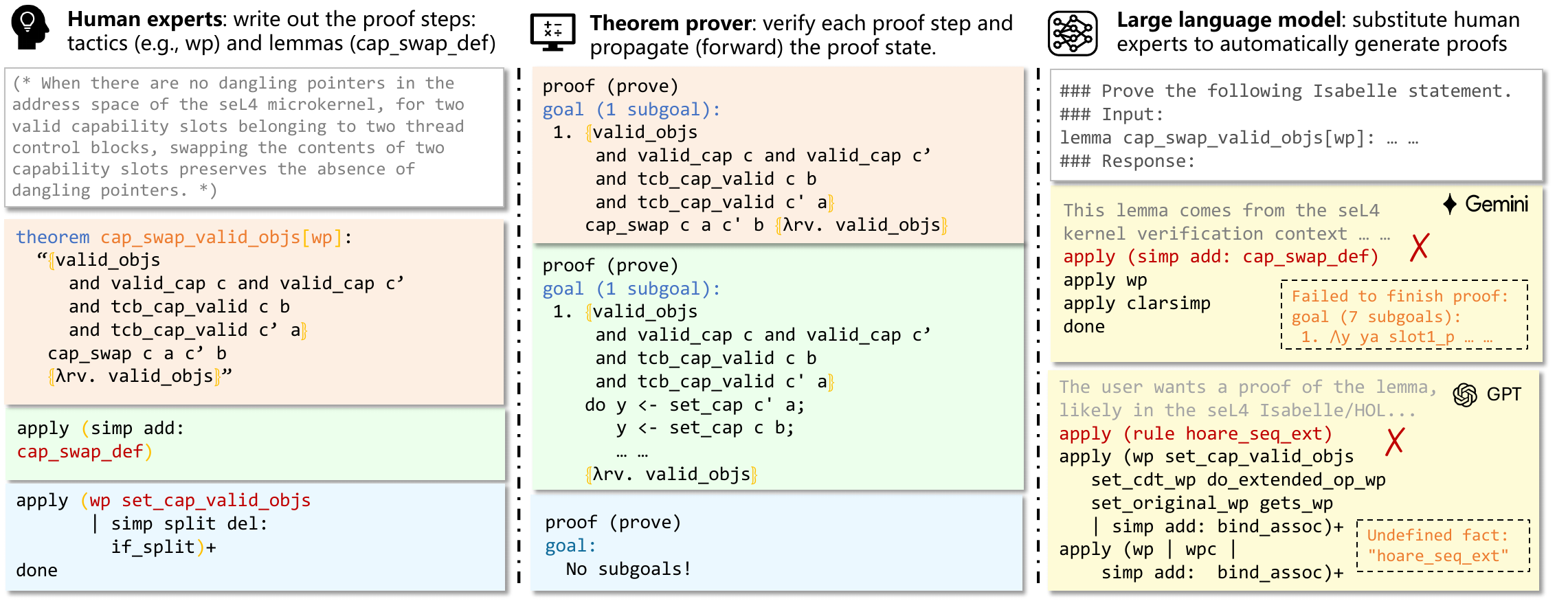}
    \caption{An example theorem and its human-written proof from the seL4 project, shown alongside LLM-generated attempts. The human-written Isabelle proof illustrates the procedural, tactic-driven reasoning style that is common in software verification. 
    We prompted two state-of-the-art LLMs, Gemini 3 Pro and GPT-5.1, to synthesize the same proof under controlled conditions, with web search disabled to avoid any leakage of the original script. Although both models correctly recognized the proof context and invoked some relevant tactics, neither succeeded in producing a valid proof even after roughly one minute of internal reasoning. Their attempts frequently misuse the domain-specific \lstinline[style=isa, breaklines=false]{wp} tactic, or hallucinate nonexistent lemmas. 
    } 
    \label{fig:example}
\end{figure*}

\smallskip
\noindent\textbf{Structure}. Section~\ref{sect:problem} illustrates the challenges of the problem. 
Section~\ref{sect:methodology} presents our neuro-symbolic approach. 
Section~\ref{sect:evaluation} reports the evaluation of our approach on seL4 and four additional repository-level benchmarks.
Section~\ref{sect:analysis} provides further qualitative analysis through a case study and an examination of failure modes.
Section~\ref{sect:related} reviews the related work.
Section~\ref{sect:lim} discusses the limitations of the current work and outlines directions for future research. 
Section~\ref{sect:conc} concludes the paper. 
\begin{figure*}[thb]
    \centering
    \includegraphics[width=0.9\textwidth]{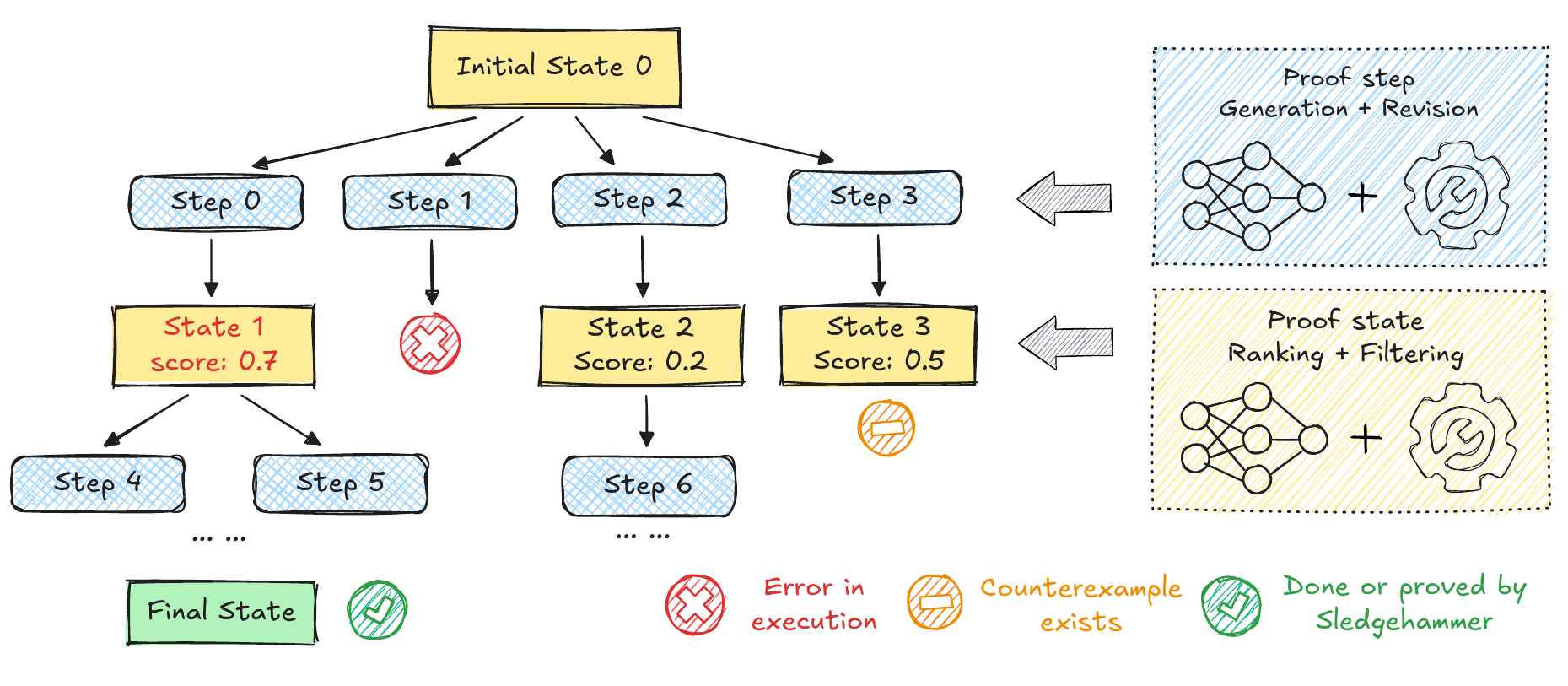}
    \caption{Overview of our neuro-symbolic proof-search framework.
    Starting from the initial proof state, the system repeatedly generates candidate proof steps using a fine-tuned LLM, followed by rule-based symbolic revision to repair syntactic or semantic issues in the proposed steps.
    Each accepted step is executed by Isabelle/HOL to derive new proof states, which are then filtered by symbolic tools to prune states containing counterexamples or duplicates. The remaining states are ranked using cumulative LLM log-probabilities, and the highest-scoring ones are selected for further expansion in a best-first search. If the search cannot complete the proof, the framework finally invokes Sledgehammer to attempt to resolve the remaining goals.}
    \label{fig:framework}
\end{figure*}

\section{Problem} \label{sect:problem}


To 
demonstrate why existing LLMs struggle to synthesize proofs in real-world software verification tasks, we present a representative theorem from seL4.
We choose seL4 as it is one of the most widely recognized formally verified systems, with its entire C implementation proven functionally correct against a formal specification in Isabelle/HOL, ensuring the absence of crashes and undefined behavior.
Building on this foundation, seL4’s verification was extended to enforce noninterference of information flows and access control, establishing formal guarantees of confidentiality and integrity.
To date, seL4 remains one of the safest OS kernels with machine-checked proofs of both implementation-level correctness and security properties, setting the benchmark for formal system verification.

An example from seL4's verification is demonstrated in Figure~\ref{fig:example}, where the \emph{theorem} \lstinline[style=isa]{cap_swap_valid_objs} is drawn from the invariant proofs of the abstract specification of the seL4 microkernel, and similar proof patterns recur throughout the project.
To prove this theorem, human experts write down (in Isabelle) three \emph{proof steps}, each comprising one \emph{tactic} (e.g., \lstinline[style=isa]{simp} and \lstinline[style=isa]{wp}) and some used \emph{lemmas} (e.g., \lstinline[style=isa]{cap_swap_def} and \lstinline[style=isa]{set_cap_valid_objs}; cf.\ the left panel in Figure~\ref{fig:example}).

Unlike formal mathematics, theorems in software verification are often formulated as Hoare triples \lstinline[style=isa]|{P} C {Q}|, 
where \lstinline[style=isa]|P| and \lstinline[style=isa]|Q| are the pre- and post-conditions, and \lstinline[style=isa]|C| denotes the program being executed.
In seL4, the abstract behavior of kernel operations is modeled in a monadic style that yields imperative-like programs; the corresponding Hoare rules capture the semantics of these monadic computations.

Intuitively, the targeted theorem states that if the address space of the seL4 microkernel contains no dangling pointers (\lstinline[style=isa]{valid_objs}), 
then swapping the contents of two valid capability slots \lstinline[style=isa]|c| and \lstinline[style=isa]|c'|, 
which belong to two thread control blocks (\lstinline[style=isa]{tcb_cap_valid c b} and \lstinline[style=isa]{tcb_cap_valid c' a}), remains dangling pointers free. 
Namely, 
after the operation \lstinline[style=isa]{cap_swap c a c' b} which exchanges the capabilities stored in these two slots, 
the resulting program state still satisfies \lstinline[style=isa]{valid_objs} expressed by the postcondition \lstinline[style=isa]{λrv. valid_objs} where \lstinline[style=isa]{rv} denotes the return value of \lstinline[style=isa]{cap_swap}.

To reduce the burden of manually constructing proofs, developers typically prepare a collection of broadly applicable tactics and lemmas, dedicated to software verification. 
For example, the \lstinline[style=isa]{wp} tactic, short for weakest precondition, generates the verification conditions associated with a given Hoare triple.
During this process, \lstinline[style=isa]{wp} systematically applies a suite of previously proved lemmas, each of which encodes a weakest-precondition characterization for a specific construct or monadic computation.
After a routine use of \lstinline[style=isa]{simp} to unfold \lstinline[style=isa]{cap_swap}, human experts can then easily invoke the pre-defined tactic \lstinline[style=isa]{wp} together with the lemma \lstinline[style=isa]{set_cap_valid_objs}.

In this example, the target theorem is discharged by combining the tactics \lstinline[style=isa]{wp} and \lstinline[style=isa]{simp}.
The simplification step performs splitting, but the usual splitting rule for \lstinline[style=isa]{if} is disabled to avoid unnecessary case distinction. 
In the proof script, the symbol \lstinline[style=isa]{|} indicates that the tactic may apply either branch (or both), while the trailing \lstinline[style=isa]{+} instructs it to repeat this process until no further progress is possible.
This interleaving enables \lstinline[style=isa]{wp} to generate obligations that \lstinline[style=isa]{simp} can reduce, while \lstinline[style=isa]{simp} reshapes the goals to facilitate subsequent \lstinline[style=isa]{wp} reasoning.
In this case, the combined step eliminates the goal entirely, and the proof concludes with \lstinline[style=isa]{done}.

To assess LLMs 
in generating this proof automatically, we experiment with two state-of-the-art proprietary models, i.e., Gemini 3 Pro~\citep{google2024gemini} and GPT-5.1~\citep{openai2025gpt51}; their results are shown on the right panel of the figure.
While both deep-thinking models successfully recognize the seL4 proof context, they are unable to apply the domain-specific tactic \lstinline[style=isa]{wp} correctly.
They also hallucinate lemmas that do not exist in the library (e.g., \lstinline[style=isa]{hoare_seq_ext}).
We further attempted several rounds of iterative repair by feeding back the ITP’s error messages, but the regenerated proofs continued to fail verification.
An analysis of the models’ reasoning traces shows that they frequently misinterpret the functionality and performance characteristics of the \lstinline[style=isa]{wp}
tactic and related lemmas (such as \lstinline[style=isa]{set_cap_valid_objs}). 

There are two potential approaches to addressing this issue.
The first is to use retrieval-augmented generation~\citep{yang2023leandojo, thompson2024rango}.
In this paradigm, given a target theorem, a retrieval module searches for relevant or structurally similar theorems and proofs, which are then supplied to the LLM as contextual guidance.
However, this strategy depends on both a high-quality proof corpus and an effective retrieval model: requirements that are difficult to satisfy in our setting.
For example, in seL4, when developers introduce a new module (such as SysInit, as shown in our evaluation) together with its accompanying proofs, comparable proofs are often absent from existing libraries.
Moreover, even when analogous theorems do exist, neither model-agnostic nor model-dependent retrieval techniques reliably identify them within a large verification codebase~\citep{blanchette2016hammering, yang2023leandojo}.
In such cases, weak retrieval can degrade, rather than improve, the quality of generated proofs.

Another approach is to fine-tune LLMs on collected proof data.
This method is considerably easier to implement and has already demonstrated promising results in formal mathematical proving.
However, fine-tuning a model to directly generate complete proofs for software verification remains highly challenging.

First, even in seL4, the available corpus consists of only about 20K theorems, which severely limits the effectiveness of fine-tuning.
Second, unlike formal mathematics, most proofs in systems like seL4 are procedural: as our example illustrates, their true semantics reside implicitly in the theorem prover’s internal proof states rather than in the proof scripts themselves.
As a result, LLMs struggle to acquire the necessary reasoning signal from surface-level proof text alone, making this approach impractical for now.
\section{Methodology} \label{sect:methodology}

We propose a neuro-symbolic approach for proof step search on system-level theorems.
The overall procedure is illustrated in Figure~\ref{fig:framework}. 
Starting from the root node, which represents the initial proof state, the search alternates between proposing new steps and deriving new proof states by executing those steps within the ITP.

To construct the next set of candidate proof steps for the current proof state, we first prompt a fine-tuned LLM to generate potential steps, and then apply a rule-based revision to refine or fix any imprecise suggestions. 
After executing these steps and obtaining a collection of new proof states, we introduce a hybrid mechanism to identify the most promising ones for further exploration.
Specifically, we apply QuickCheck~\citep{claessen2000quickcheck} and Nitpick~\citep{blanchette2010nitpick}, which serve as counterexample generators that help detect false conjectures or invalid intermediate proof states in Isabelle/HOL, to discard any proof state that contains a counterexample. 
We then use the LLM’s prediction score (cumulative perplexity) to rank the remaining states and select the most promising one.
Finally, if the search for the proof fails, we invoke an automated tool, i.e., Sledgehammer, to attempt to complete the proof. 
Putting all the above together, we establish the framework in Algorithm~\ref{alg:best_first}.

\begin{algorithm}[t]
\caption{Proof Search Framework}
\label{alg:best_first}
\begin{algorithmic}[1]

\Require initial state $s_0$, neural generator $\mathcal{G}$, symbolic checker $\mathcal{C}$, max iterations $I_{\max}$
\State initialize tree $T$ with root containing $s_0$
\For{each iteration $i \in [1, I_{\max}]$}
    \State $n \gets$ highest\_score\_unexplored($T$) 
    \State succStates $\gets []$; failStates $\gets []$ 
    \State candidates $\gets \mathcal{G}(n.state)$ \hfill {\color{deepgreen}$\triangleright$ step generation}

    \For{each step $\delta \in$ candidates}
        \State $s \gets$ apply($\delta$, $n.state$) \hfill {\color{deepgreen}$\triangleright$ ITP execute}
        \If{$s$ is error state} 
            \State failStates.add($(s, \delta, \log p_{\delta})$)
        \Else
            \State succStates.add($(s,\delta, \log p_{\delta})$) 
        \EndIf
    \EndFor

    \For{each $(s, \delta, \log p_\delta)$ in \textsc{Revise}(failStates)}
       \State $s \gets$ apply($\delta$, $n.state$)
       \State succStates.add($(s, \delta, \log p_\delta)$)
    \EndFor

    \For{each $(s,\delta, \log p_\delta)$ in succStates}
        \If{s has no subgoals} 
            \State \Return proof script \hfill {\color{deepgreen}$\triangleright$ proof completed}
        \EndIf

        \If{$\mathcal{C}(s)$ detects duplicate or counterexample} 
            \State \textbf{continue} \hfill {\color{deepgreen}$\triangleright$  state filtering}
        \EndIf

        \State score $\gets$ combine($n.score$, $\log p_{\delta}$) \hfill {\color{deepgreen}$\triangleright$ scoring}
        \State insert\_node($T$, $(s, score)$) 
    \EndFor
\EndFor

\State \Return \textsf{FAIL}
\end{algorithmic}
\end{algorithm}


\subsection{Proof Step Generation \& Revision}
\subsubsection{Proof Step Generation} 
Our approach begins by generating candidate proof steps through prompting a fine-tuned LLM.
Given the current proof state, which encapsulates the available hypotheses and the target goal, we query the model to propose the next proof step using a simple instruction template:
\begin{boxA}
{\fontsize{8}{8.8}\selectfont
\begin{verbatim}
### Given the following Isabelle proof state,
suggest the next proof step.
### Input:
<current_proof_state>
### Response:
<next_proof_step>
\end{verbatim}
}
\end{boxA}

The rationale for 
focusing on the current proof state is twofold.
First, automated theorem proving can be cast as a sequential decision-making problem, and thus can be modeled as a Markov decision process, in which a standard result stipulates that pure, history-independent policies (i.e., the next step depends solely on the present state) suffice for the reachability objective, i.e., to reach the ``proof done'' state. This is also a common practice in the literature of proof search, which turns out to be critical for efficiency~\citep{wu2021tacticzero, huang2025leanprogress, wu2020neural}.
Second, the structure and vocabulary of the proof state provide strong cues regarding which tactics 
are applicable and which premises are likely to be relevant~\citep{gauthier2021tactictoe, piotrowski2023machine}.
Additionally, we focus on predicting a single step at a time. 
This is because, although multi-step prediction is possible in principle, the complexity of real proofs and the limited amount of available data make such predictions considerably more challenging; empirically, one-step prediction yields more reliable results.

To fine-tune the model to predict such next steps, we construct a dataset of state–step pairs extracted from existing theorems and their complete proofs.
Each proof is decomposed into individual proof steps and replayed in the ITP, allowing us to record the intermediate proof state preceding every step.
Using these aligned pairs, we fine-tune the LLM to generate outputs that are not only syntactically correct but also semantically sensible.
Finally, since proof search demands efficient generation, we employ relatively small LLMs (e.g., 1.7B or 7B size), which are nevertheless sufficient for our specific tasks.

\subsubsection{Proof Step Revision}

Writing syntactically and semantically valid proof steps in an ITP (such as Isabelle) is challenging, particularly in system verification tasks that depend on numerous domain-specific lemmas and tactics. 
Consequently, even a well-trained LLM may generate steps that either refer to premises or tactics absent from the current proof context, or cannot be applied to the current goals due to syntactic errors or lack of progress. 
Such failures substantially degrade the effectiveness and efficiency of the proof search.

Despite their incorrectness, these failed attempts often contain useful information. Continuing our running example, an LLM might identify relevant premises {\lstinline[style=isa]{cap_swap_def}}
but pair them with an inapplicable tactic \lstinline[style=isa]{by}; 
or it may propose a promising tactic \lstinline[style=isa]{wp} 
while selecting premises \lstinline[style=isa]{hoare_seq_ext} that do not precisely fit the context.

To fully exploit these incorrect but reasonable responses from LLMs, we employ a best-effort repair procedure. 
Whenever the ITP rejects an LLM-generated step, we record it for post-processing and re-evaluate the revised version.
After examining all proposed steps for a given proof state, we systematically transform the failed ones in two ways: 
(1) tactic repair: we extract the premises referenced by misused tactics and recombine them with a curated set of frequently used tactics; and 
(2) premise repair: we search the proof context for similar premises and substitute them where appropriate.
By synthesizing additional candidate steps, these symbolic revisions help reveal prover-accepted steps, thereby strengthening the overall proof search.

\begin{algorithm}[t]
\caption{Proof-step revision and execution}
\label{alg:correction}
\begin{algorithmic}[1]
\Require Failed proof state set failStates

\State revisedSteps $\gets []$
\State premiseSet $\gets$ established from trainingCorpora 
\State tacticSet $\gets$ established from trainingCorpora
\For{each $(s, \delta, \log p_\delta) \in$ failStates}

    \If{$s$.error is w.r.t. tactic}  \hfill {\color{deepgreen} $\triangleright$ tactic repair}
    \State $p$ $\gets$ extract referred premises in $\delta$
    \State revisedSteps.add(combine(tacticSet, p))
    \EndIf
    
    \If{$s$.error is w.r.t. premise} \hfill {\color{deepgreen}$\triangleright$ premise repair}
    \State $t/p \gets$ extract referred tactic/premises in $\delta$
    \State $p' \gets$ retrieve similar premises from premiseSet
    \State revisedSteps.add(combine($t$, $p'$))
    \EndIf
    
\EndFor

\State \Return revisedSteps
\end{algorithmic}
\end{algorithm}

For the tactic repair, we first extract all tactics from the training corpus. 
Given a proof step rejected by the ITP, we attempt to rewrite it by identifying the premises referenced in the failed step and recombining them with the pre-built tactic set.
As to the premise repair, which explicitly targets proof steps that fail with an ``undefined fact'' error, we attempt to correct every LLM-generated undefined premise using an available version from the proof context.
Concretely, we compare the undefined fact with all candidate premises using edit distance and select the closest matches (e.g., the top three). The failed step is then revised by substituting the matched premises for the undefined ones.

The procedure of proof-step revision is summarized in Algorithm~\ref{alg:correction}. 
However, the revision can produce a large number of candidate steps. 
To alleviate this, we impose constraints to keep the search space manageable. 
For tactic repair, we limit the tactic set (e.g., to the 12 most frequently used tactics).
For premise repair, we pre-select a compact subset of premises from the library.
In particular, we use MePo (a heuristic module in Isabelle’s Sledgehammer that ranks and selects premises relevant to a given goal) as a lightweight relevance filter to retrieve the 128 facts most related to the current subgoal. 


\subsection{Proof State Filtering \& Ranking}

\subsubsection{Proof State Filtering} \label{sect:psf}
To alleviate state explosion in tree-search approaches, we reorganize the accumulated proof states before further exploration. 
Particularly, after generating new proof states by executing the newly produced proof steps, we first invoke two complementary symbolic tools, QuickCheck~\citep{claessen2000quickcheck} and Nitpick~\citep{blanchette2010nitpick}, to detect potential counterexamples in the candidate states. QuickCheck performs random property-based testing on executable parts of the proof state, whereas Nitpick translates higher-order formulas into finite relational models and searches for counterexamples using Kodkod, which is a SAT-based first-order relational model finder. 

Additionally, we adapt the tool SolveDirect~\citep{SolveDirect2025} to detect potentially duplicate proof states.
Initially, SolveDirect determines whether a newly stated theorem can be solved directly using an existing one.
In our setting, we formalize two proof states as theorems and apply SolveDirect in both directions to determine whether they are semantically equivalent.

We use a basic 
theorem in seL4 as an example to illustrate 
proof state filtering. 
Consider the ``signed overflow'' theorem, whose formal statement is as follows.

\begin{isacode-blue}
theorem sofl_test:
  ‹ sint x + sint y = sint (x + y) ⟷
      drop_bit (size x - 1)
          ((x + y XOR x) AND (x + y XOR y)) = 0 ›
  for x y :: ‹ 'a::len word ›
\end{isacode-blue}
The theorem claims that for two signed machine words, the value of their machine-level sum matches the sum of the mathematical values they represent. This property guarantees the correctness of the machine-word definitions on which the low-level kernel specifications depend.
To prove this theorem, the LLM proposes a total of 8,445 candidate proof steps, rendering exhaustive exploration prohibitively time-consuming.
Nevertheless, our filtering reveals that 44.2\% of these steps lead to duplicate proof states, and 52.3\% of the remaining states are unprovable due to existing counterexamples.
Thus, the state-filtering process effectively eliminates unproductive effort and conserves the search budget, enabling the proof to be found much earlier.

\subsubsection{Proof State Ranking}

To guide the proof search efficiently, we rank all proof states and always expand the most promising ones first.
To score a given proof state, we rely on the log probabilities predicted by the fine-tuned LLM.
The LLM is trained to predict the next proof step; its log probability directly reflects the model’s confidence in applying that step to the current proof state.

Consider a candidate proof state $s_L$, obtained by applying a sequence of proof steps $(a_1, \dots, a_{L-1})$.
We define its cumulative log-probability score as
\[
\text{score}(s_L)
= \frac{\sum_{t=0}^{L-1} \log p(a_t \mid s_t)}{L^{\alpha}},
\]
where $s_1, \dots, s_{L-1}$ are the intermediate states produced by the corresponding steps. The normalization term $L^{\alpha}$ mitigates the inherent bias against longer proof sequences, with the exponent $\alpha$ specifying the degree of mitigation~\cite{xin2025bfs}, which we set to 1 by default. After computing the score for each proof state, we select the top-\textit{k} states for further expansion where $k$ is a hyperparameter.

\subsection{Hammer Integration}
Hammers are tools designed to discharge proof goals in ITPs automatically. 
In Isabelle, the primary hammer is Sledgehammer~\cite{bohme2010sledgehammer}, which applies ATPs and SMT solvers (e.g., Z3, CVC5, E, SPASS, Vampire, etc) to the current goal and synthesizes corresponding Isabelle proofs.
When invoked, it first retrieves a fixed number of potentially relevant premises that may aid in solving the goal.
These premises, together with the goal, are translated into a form suitable for backend ATP and SMT solvers. If a backend solver succeeds, Sledgehammer attempts to reconstruct a corresponding Isabelle proof step, and, when this reconstruction succeeds, a complete proof for the current goal is returned to the prover.

Although Sledgehammer’s performance is far from satisfactory for our tasks, it plays a mutually beneficial role in the tree search process.
In particular, while tree search may fail to solve the original problem within a limited search budget, it often reduces the problem to a simpler form that Sledgehammer can solve, even though it cannot solve the initial goal on its own. This makes it natural to integrate the hammer into our framework. Because invoking the hammer at every search step is prohibitively expensive, we only call it when tree search fails.
Specifically, upon search failure, we rank all proof states in the currently expanded tree, select a small subset with the highest scores, and pass them to Sledgehammer for a final attempt at solving the goal.

\section{Evaluation} \label{sect:evaluation}

This section presents our experimental results. 
We use seL4 as a system-level evaluation testbed to demonstrate both the effectiveness and efficiency of our proposed method. 
In addition, we introduce four additional repository-level benchmarks to further assess its generalizability.
Specifically, our study aims to address the following research questions (\textbf{RQ}s):
\begin{itemize}[leftmargin=1em]
\item \textbf{RQ1 – Effectiveness}: Does our proposed method outperform existing techniques in generating proofs for seL4?
\item \textbf{RQ2 – Efficiency}: How much human effort in proving seL4 can be saved using our proposed method?
\item \textbf{RQ3 – Generalizability}: Can our method perform well on the verification of other software projects?
\item \textbf{RQ4 – Ablation}: What is the contribution of each 
component within our framework?
\end{itemize}
All evaluations were carried out within a dedicated Docker environment on a Linux server with an AMD EPYC 9654 96-core processor and six high-end GPUs. 

\subsection{Isabelle REPL}

To enable efficient interaction with Isabelle from Python, we implement a new Isabelle REPL (Read-Eval-Print Loop), whose logic builds on scala-isabelle~\cite{unruh2025scalaisabelle} and Py4J~\citep{nishant2023py4j}.
The REPL provides a Gateway Server that exposes Isabelle components to Python clients through a unified API. 
This design enables programmatic interaction with Isabelle, allowing users to parse Isabelle theories, extract run-time context information, and execute proof steps incrementally.

Compared to the previously developed Isabelle REPL, PISA~\cite{jiang2021lisa}, our REPL supports more recent Isabelle versions and provides extended ML-level capabilities.
A wide range of Isabelle’s proof-automation tools, including Sledgehammer, Nitpick, QuickCheck, and other integrated methods, can be accessed directly through our interface.
It also improves the extraction of information from proof contexts, such as variables and assumptions in proof goals, dependencies of proved theorems, and facts selected by Sledgehammer.

In addition, our REPL provides improved management of the Isabelle process, top-level state, and theories. 
It allows users to safely impose strict time limits on Isabelle processes, as well as to clone, restore, and switch between different proof states.
Theory management is also strengthened: in large verification projects, complex theory dependencies can heavily degrade compilation efficiency. 
For example, in the CRefine theory of seL4, a single proof step takes more than one minute to process.
To address this, we implement a caching mechanism that avoids redundant execution of unrelated theorems and proofs, significantly reducing unnecessary recomputation.

\subsection{Experimental Setup}

\noindent\textbf{Dataset.} 
We use the FVELER dataset~\citep{lin2024fvel}, which comprises 29,125 theorems from seL4. The dataset is split into four subsets, i.e., training, validation, test, and test-hard, containing 26,081, 1,115, 1,077, and 852 theorems, respectively. Following the FVEL criteria, the training, validation, and test sets are randomly partitioned. 
In contrast, the test-hard set consists of theorems drawn from three specific and independent sessions, i.e., SysInit, SysInitExamples, and LibTest, that do not appear in any of the other splits. Note that \textit{sessions} are the way Isabelle organizes theory files, which resemble the relationship between code files and libraries in other programming languages. A theory can import other theories, and a session can depend on other sessions by declaring dependencies in the Isabelle ROOT file.
These sessions are typically selected based on the depth of their session-level dependency graphs, and the theorems they contain exhibit more complex dependencies, making them considerably more challenging to prove.

In our experiments, we use the training set as the corpus for fine-tuning LLMs, and adopt validation, test, and test-hard sets to evaluate our proposed framework.
From the training set, we further extract all proof state–step pairs, yielding a total of 181,887 pairs.
On average, each theorem contributes to about seven pairs, substantially increasing the amount of supervised data available for model training.


\smallskip
\noindent\textbf{Training.} We fine-tuned two small models, Qwen3-1.7B~\citep{yang2025qwen3} and Mistral-7B~\citep{jiang2023mistral}, to predict and score each proof step.
We carried out the full-parameter supervised fine-tuning (SFT) using the Llama-Factory framework~\citep{zheng2024llamafactory}.  
We use DeepSpeed ZeRO-2 optimization to enable memory-efficient distributed training. 
The training used an effective global batch size of 16 and ran 3 epochs using bfloat16 mixed precision. The learning rate was set to 1e-5 and followed a cosine decay schedule with a warmup ratio of 0.1. 

\smallskip
\noindent\textbf{Proving.}
For the LLM inference, we adopt a high temperature of 1.0 and a top-\textit{p} value of 0.95 to explore the proof space sufficiently. 
The maximum generation length is fixed at 2,048 tokens.
In each search iteration, at most five proof states are selected; for each proof state, the model is prompted to generate 128 candidate proof steps.
In our framework, we disable LLMs' internal thinking to improve inference efficiency. 

When proof search fails, we select the 16 highest-scoring proof states and invoke Sledgehammer for assistance. We run Sledgehammer with its default premise-selection strategy: it first selects 2,048 most relevant premises for the current proof goal using MeSh, which combines MePo~\cite{meng2009lightweight} and MaSh~\cite{kuhlwein2013mash} to gauge relevance between the proof goal and available facts in the context.
With these retrieved results, Sledgehammer then attempts to discharge the current subgoal using the built-in SMT/ATP solvers Z3, CVC5, E, SPASS, and Vampire, with a 60s time limit.

\smallskip
\noindent \textbf{Baseline.}
We select four neural or symbolic proof-generation methods as baselines.
Selene prompts GPT-4o with the target problem and several examples to generate a candidate proof.
FVEL follows a similar paradigm but also fine-tunes a Mistral-7B-Instruct model on the training data to produce complete proofs, rather than using in-context learning.
We also include two symbolic methods, Auto and Sledgehammer, which operate purely symbolically to prove the target theorem.
Here, Auto is a collection of common symbolic automatic proving methods in Isabelle, such as \lstinline[style=isa]{by simp}, \lstinline[style=isa]{by auto}, and \lstinline[style=isa]{by fastforce}.
The time limit for all methods is set to 120 minutes, as we observed no noticeable improvement when increasing it further.

\subsection{Experimental Results}

\subsubsection{RQ1: Effectiveness Comparison}


\begin{table}[t]
\centering\small
\setlength{\tabcolsep}{2.5pt}
\caption{Proof success rates (\%) across validation (Val), test (Test), and test-hard (TsHd) splits. Compared to existing neural and symbolic provers, our framework delivers substantial improvements: Qwen3-1.7B achieves 70.4\% overall success, and Mistral-7B reaches 77.6\%, exceeding the strongest baseline by 30.0 and 37.3 percentage points, respectively.}
\label{tab:results}
\begin{tabularx}{\columnwidth}{ll*{4}{>{\centering\arraybackslash}X}}
\toprule
\multicolumn{2}{c}{\textbf{Method}} & \textbf{Val} & \textbf{Test} & \textbf{TsHd} & \textbf{Total} \\
\midrule
\multirow{2}*{Neural} & Selene  & 6.1 & 7.0 & 3.3 & 5.6 \\
& FVEL & 8.9 & 9.5 & 4.5 & 7.8 \\
\midrule
\multirow{2}*{Symbolic} & Auto & 4.9 & 6.7 & 6.1 & 5.9 \\
& Hammer & 40.5 & 39.5 & 40.9 & 40.3 \\
\midrule
\multirow{4}*{Ours} & Qwen3 & {73.6} & {74.9} & {61.1} & {70.4} \\
& \cellcolor[gray]{.92} $\Delta$ & \cellcolor[gray]{.92} $\uparrow$33.1 & \cellcolor[gray]{.92} $\uparrow${35.4} & \cellcolor[gray]{.92} $\uparrow$20.2 & \cellcolor[gray]{.92} $\uparrow$30.0 \\
& Mistral & \textbf{79.8} & \textbf{89.0} & \textbf{69.8} & \textbf{77.6} \\
& \cellcolor[gray]{.92} $\Delta$ & \cellcolor[gray]{.92} $\uparrow$\textbf{39.3} & \cellcolor[gray]{.92} $\uparrow$\textbf{49.5} & \cellcolor[gray]{.92} $\uparrow$\textbf{28.9} & \cellcolor[gray]{.92} $\uparrow$\textbf{37.3} \\
\bottomrule
\end{tabularx}
\end{table}

Table~\ref{tab:results} summarizes the performance of our approach compared to both neural and symbolic baselines. 
Among neural methods, Selene achieved 156 successful proofs, while FVEL improved this to 219.
As for symbolic baselines, Auto proved 164 theorems, whereas Sledgehammer reached 1,124. 
In contrast, our framework substantially outperforms all these baselines across different splits.
Using the fine-tuned Mistral-7B model, our approach produces a total of 2,167 successful proofs, of which 788 are on the validation set, 811 are on the test set, and 568 are on the test-hard set.
This accounts for 77.6\% of all still valid theorems across the three evaluated sets, 37.3 percentage points higher than the best baseline, Sledgehammer.
It is also encouraging that, on a relatively small model, Qwen3-1.7B, our approach manages to complete the proof of 1,965 theorems, which amounts to 70.4\% of the total theorems. 
This demonstrates that in a real environment where new proofs need to be crafted, a deployed small model can already help with the development of verification.

\begin{figure}[t]
\centering
\includegraphics[width=0.47\textwidth]{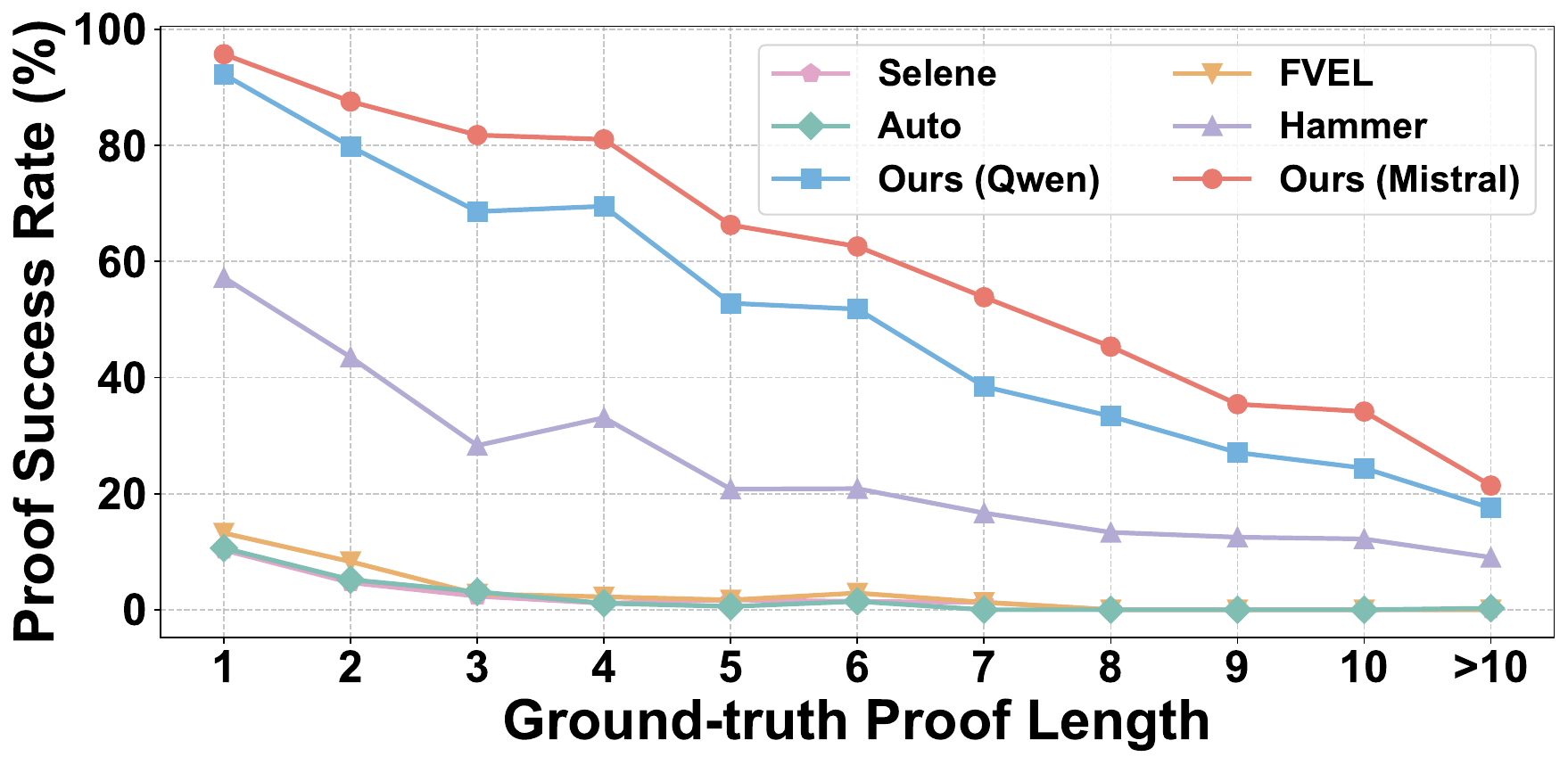}
\caption{Proof success rate across ground-truth proof lengths. Our approach consistently outperforms existing methods. In particular, the success rate decreases as proofs grow longer, but our approach still maintains non-negligible performance even on proofs exceeding 10 lines.}
\label{fig: correctness with length}
\end{figure}

We further evaluate the performance of our automated proving approach in two aspects. First, prior work struggled to generate nontrivial proofs. Since the proofs in seL4 are largely procedural, we use the proof length as a proxy for proving difficulty. 
We compute the proof success rate across different proof lengths and present the results in Figure~\ref{fig: correctness with length}.
The results show that our approach is more capable of proving longer theorems than previous baselines. 
Although the success rate naturally decreases as proof length increases, it remains around 20\% even for the 393 theorems whose proofs exceed 10 lines. 
We leave it as future work to investigate more effective strategies for constructing longer proofs.

\begin{figure}[t]
\centering
\includegraphics[width=0.47\textwidth]{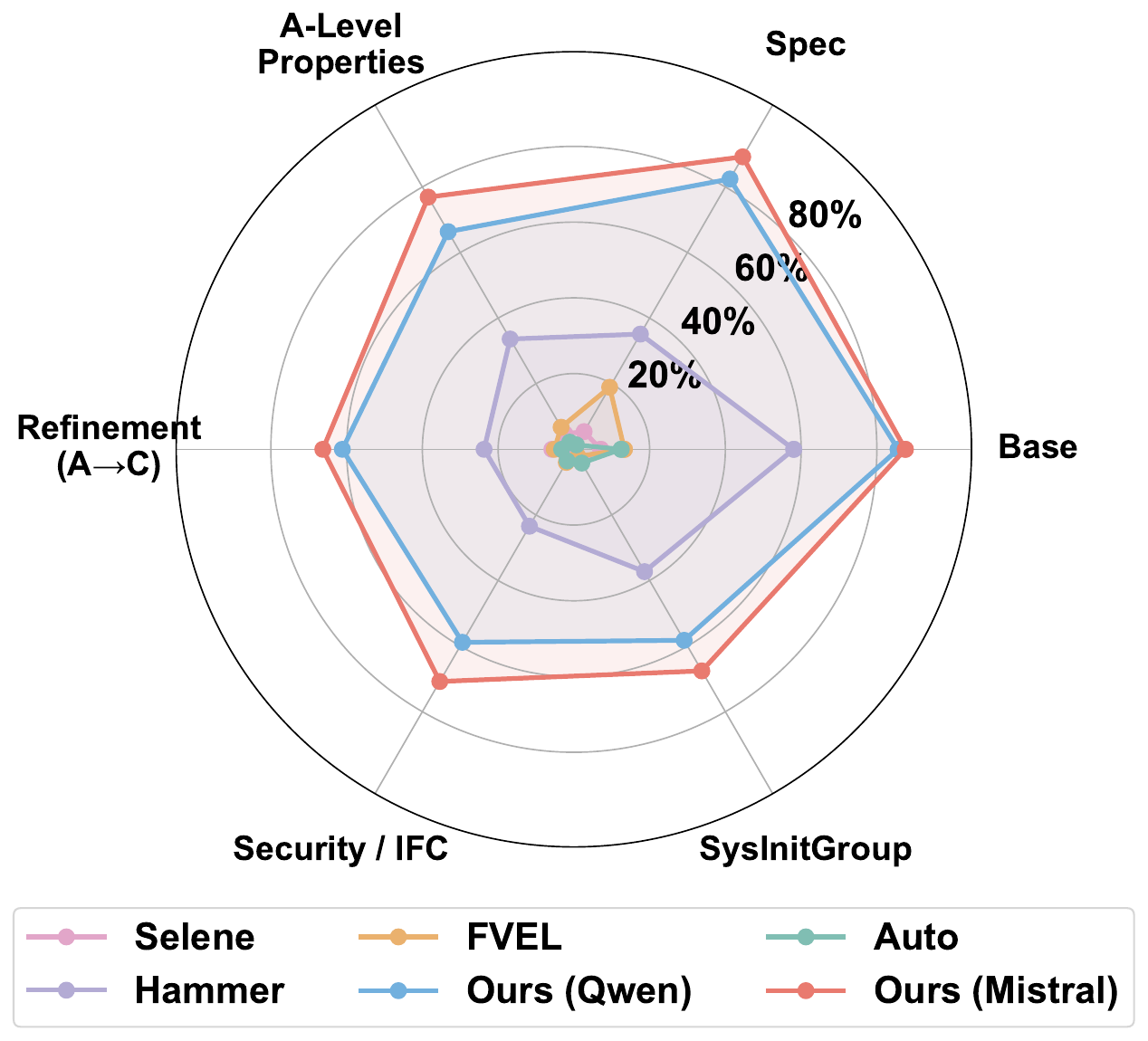}
\caption{Proof success rates across the six seL4 session categories. Our approach yields stronger performance than all baselines in all sessions. Notably, it achieves high accuracy even on the challenging SysInitGroup category, which is entirely unseen during training.}
\label{fig:correctness with sessions}
\end{figure}

Second, we examine the performance across different proof sessions.
We group the seL4 theorems into six major categories w.r.t. the functionality~\citep{seL4Integrity, seL4InfoFlow, klein2014comprehensive}: Base Libraries \& Tools (Base), Specifications (Spec), Abstract-Level (A-Level) Properties, Refinement from Abstract-Level to Executable C  (A$\to$C), Security/Information Flow Control (IFC), and System Initialization Group (SysInitGroup).
Figure~\ref{fig:correctness with sessions} presents the results across these categories, demonstrating that our approach delivers superior performance compared with all baselines across all sessions.
Notably, under the test-hard set division strategy, SysInitGroup is derived from a completely distinct session and is isolated from the training data, yet our approach still achieves a success rate of 67.6\%.
 
\begin{tcolorbox}[
  colback=gray!5,
  colframe=black!60,
  title=Response to RQ1
]
Our approach substantially outperforms baselines, achieving a success rate 37.3 percentage points higher than the (symbolic) ATP tool Sledgehammer and 69.8 percentage points higher than the state-of-the-art neural method.
It remains reliably effective across different proof difficulty levels and session categories.
Further, it demonstrates good generalizability, 
successfully discharging 67.6\% of the theorems from unseen sessions during training.
\end{tcolorbox}

\begin{table}[t]
\centering\small
\setlength{\tabcolsep}{2.5pt}
\caption{Coverage results of automatically generated proof lines. Our method attains the highest coverage, proving theorems whose existing proofs amount to over one-third of all proof lines and substantially outperforming existing baselines in reducing manual proof-script development effort.}
\label{tab:line of proofs}
\begin{tabular*}{\columnwidth}{l@{\extracolsep{\fill}}rr}
\toprule
\textbf{Method} & \textbf{\# of proof lines} & \textbf{\% of coverage} \\
\midrule
Selene & 197 & 1.1 \\
FVEL & 271 & 1.6 \\
Auto & 204 & 1.2 \\
Hammer & 2,581 & 15.0 \\
\midrule
\textbf{Ours} & \textbf{6,235} & \textbf{36.2} \\
\bottomrule
\end{tabular*}
\end{table}

\begin{figure}[htb]
\centering
\includegraphics[width=0.47\textwidth]{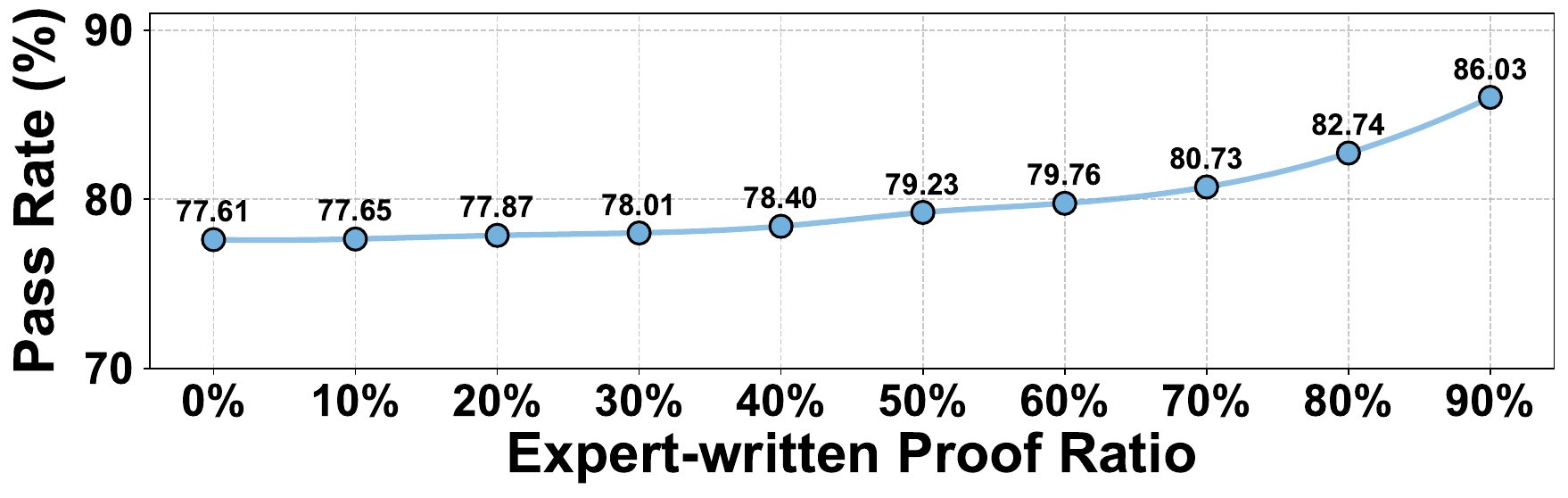}
\caption{Proof completion rate under varying proportions of expert-provided proof lines. Our method is increasingly effective as more human-written lines are supplied. Overall, the method reduces human effort by 71.1\% on average.}
\label{fig: completion curve}
\end{figure}


\subsubsection{RQ2: Proving Efficiency in Saving Human Efforts}

To evaluate how effectively our approach reduces manual effort in developing proof scripts, we adopt a standard metric: the coverage of the entire library that can be automatically generated~\cite{matichuk2015cost}. As summarized in Table~\ref{tab:line of proofs}, our method successfully proves theorems whose corresponding proofs account for 6,235 lines in the seL4 proof corpus, achieving 36.2\% coverage across all evaluated theorems. Under this metric, our approach yields more than a one-fold improvement over existing baseline methods.

To evaluate how effectively our approach reduces manual proof effort, we simulate an AI--human collaborative workflow: for each theorem, we supply a prefix of the ground-truth proof (the first $\sigma$ fraction of expert-written lines) and ask our approach to complete the remaining steps. Figure~\ref{fig: completion curve} plots the proof completion rate as a function of the prefix ratio $\sigma$.

As the prefix grows, more theorems become completable. Including cases solved entirely automatically ($\sigma=0$), our approach saves at least 10\% of the expert proof effort for 86.0\% of the theorems. To summarize this curve with a single metric, we define the average effort saving $\eta$:
\begin{equation*}
\eta = \sum_{i=1}^{k} (p_i - p_{i-1}) \cdot \sigma_i, \quad p_0 = 0,
\end{equation*}
where $(\sigma_1, \dots, \sigma_k)$ are the prefix ratios in ascending order, and $p_i$ is the fraction of theorems whose remaining proof can be completed automatically given a $\sigma_i$-prefix. This yields $\eta = 71.1\%$, meaning our approach reduces human proof effort by roughly two-thirds on average.

We also record the time consumption of our framework as a reference; the average time to generate a successful proof is 139.1 minutes. 
However, it is worth noting that this average is skewed upward by a small number of theorems that take extremely long to prove. 
In our experiment, we find that 58.4\% of the theorems are proved within 10 minutes, 73.4\% within 30 minutes, and 80.8\% within 2 hours.
Therefore, the majority of proofs are obtained within minutes to tens of minutes---substantially faster than manual proof development, which often requires hours to days per theorem for complex system properties.

\begin{tcolorbox}[
  colback=gray!5,
  colframe=black!60,
  title=Response to RQ2
]
Our approach alleviates much of the burden of interactive theorem proving, automatically completing proofs that correspond to 36.2\% of the proof lines in the evaluated corpus and reducing expert effort by approximately 71.1\% in an AI–human collaboration setting. Moreover, it is able to synthesize the majority of proofs within a practical time budget.
\end{tcolorbox}


\subsubsection{RQ3: Generalizability to Other Projects}

A major concern with training-based methods is their generalizability, namely, whether the approach is only applicable to a specific domain due to potential data leakage or overfitting. 

\begin{figure}[t]
\centering
\includegraphics[width=0.47\textwidth]{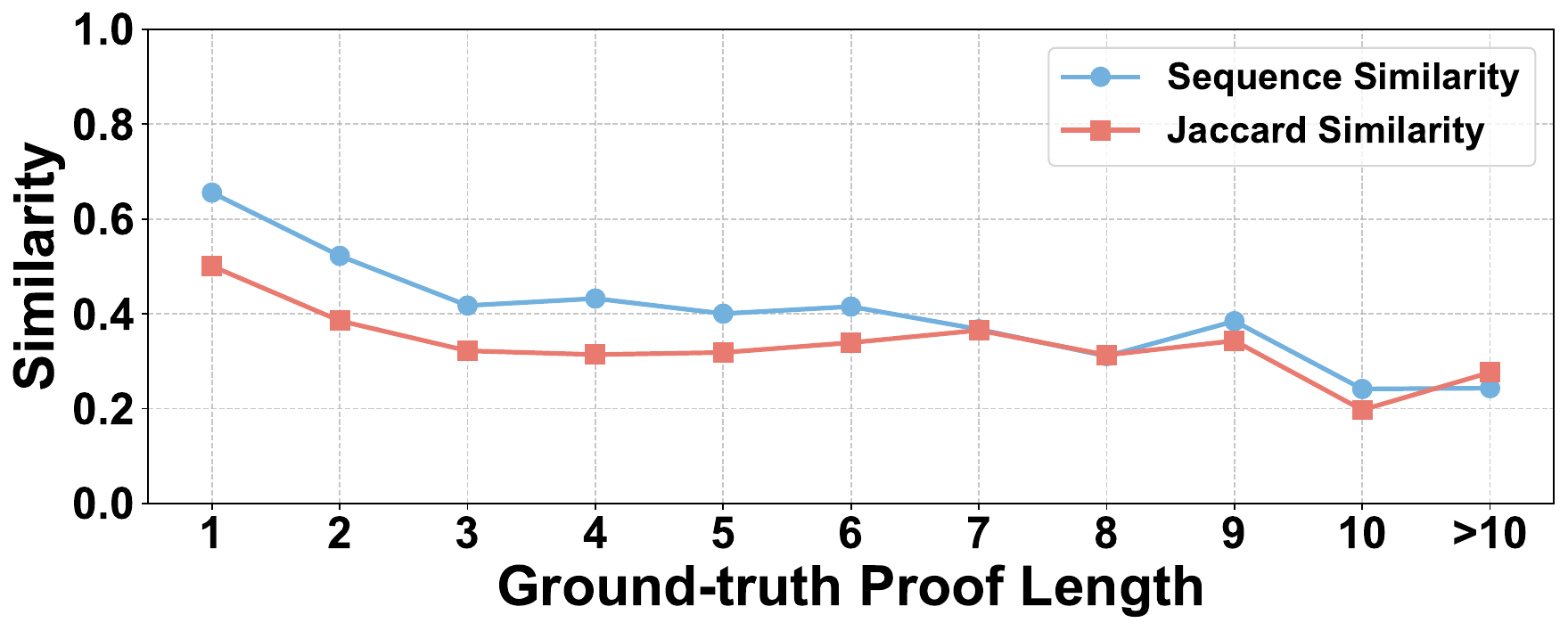}
\caption{
Sequence and Jaccard similarity between generated and ground-truth proofs across proof lengths. Both metrics remain low and decrease as proofs grow longer, suggesting the model synthesizes novel strategies rather than recalling memorized proofs.
}
\label{fig:similarity curves}
\end{figure}

To investigate possible leakage, given that the seL4 proofs are publicly available and that the pre-trained model may have already incorporated some of the knowledge, we compute the similarity between the ground-truth proofs and the proofs generated by our approach using sequence similarity~\citep{levenshtein1966binary} and Jaccard similarity~\citep{jaccard1901etude}. As shown in Figure~\ref{fig:similarity curves}, both metrics remain low and, crucially, decrease as proof length grows. If the model were memorizing ground-truth proofs, we would expect similarity to remain high regardless of length; the observed downward trend instead suggests the model is synthesizing novel proof strategies rather than recalling memorized ones. The only exceptions are one- or two-line proofs, where the solution space is inherently small---yet even among these, only 16.7\% are identical to the ground truth.

We further explore the generalizability of our approach and the fine-tuned LLM.
We select three projects from the Archive of Formal Proofs (AFP\footnote{\url{https://www.isa-afp.org/}}), i.e., X86 Semantics~\cite{X86_Semantics-AFP}, IEEE Floating Point~\cite{IEEE_Floating_Point-AFP}, and SATSolverVerification~\cite{SATSolverVerification-AFP}, to directly apply proof search on the theorems therein. In addition, we include the code verification benchmark Code2Inv and translate its loop invariants into Isabelle theorems.

\begin{figure}[t]
\centering
\includegraphics[width=0.45\textwidth]{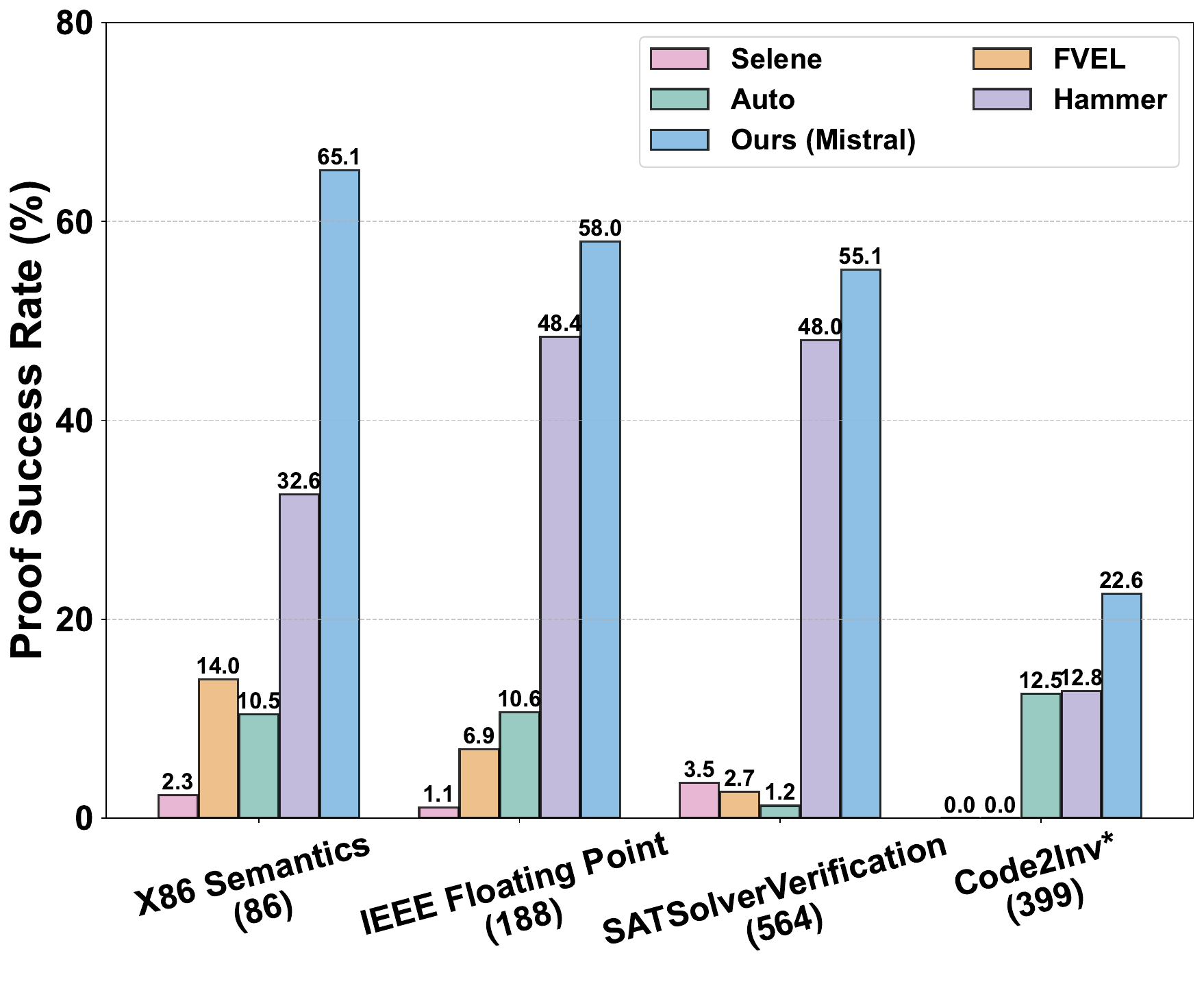}
\caption{Performance of our proof search framework on additional benchmarks. 
The $x$-axis lists the benchmarks, where entries in brackets denote the number of theorems included in each benchmark. 
Our approach consistently achieves higher proof success rates across all benchmarks, demonstrating strong generalizability and robustness.}
\label{fig:afp results}
\end{figure}

The performance of our approach and baselines is shown in Figure~\ref{fig:afp results}. 
Because the X86 semantics benchmark is more relevant to the seL4 project, the fine-tuned LLM performs particularly well on it, yielding a 32.5\% higher success rate than the baselines.
The other three benchmarks are hammer-friendly because of their SMT-oriented nature. Even in this setting, our framework demonstrates superior effectiveness, achieving an average relative improvement of 36.9\%.

\begin{tcolorbox}[
  colback=gray!5,
  colframe=black!60,
  title=Response to RQ3
]
The knowledge acquired by the fine-tuned LLM, together with our proof search framework, generalizes well across domains. In particular, the approach can be effectively leveraged to assist proof construction in a wide range of formal verification tasks beyond the seL4 setting.

\end{tcolorbox}

\subsubsection{RQ4: Ablation study}

Since our approach integrates several components, namely LLMs, hammer, as well as proof-step revision and proof-state filtering mechanisms, we conduct ablation studies to evaluate the contribution of each part of the framework.
To assess the impact of our fine-tuned model, we replace it with off-the-shelf LLMs such as GPT-4o or DeepSeek.
Due to budget constraints (running tree search costs approximately \$4.1 per theorem on average when using the DeepSeek or GPT-4o APIs), we randomly sample 200 theorems from the benchmark and use the DeepSeek API to propose proof steps.

In addition, inspired by Rango~\citep{thompson2024rango}, we implement a retrieval-augmented generation (RAG) mechanism.
The retrieval module fetches both similar proof steps and potentially relevant premises.
Similar steps are retrieved from the training set based on the similarity between their associated proof states and the current proof state, where these states are tokenized into sequences of identifiers, and relevance is measured using BM25~\citep{robertson2009probabilistic}.
For premises retrieval, we continue to rely on Sledgehammer’s MePo heuristic for efficiency.
The final prompt for LLMs includes the retrieved steps and premises along with their statements, and we restrict the total lengths of steps and premises to 1024 and 512 tokens, respectively.

The results for all method variations are reported in Table~\ref{tab:api test}.
Overall, although adding tree search and RAG steadily improves DeepSeek's performance, these variants still fall significantly short of our full framework.
This confirms that the combination of the fine-tuned model, structured search, and hammer integration is essential for achieving high proof success rates, and highlights the importance of each major component in our design.

We also evaluate the impact of the fine-tuning process by adopting a hammer-free tree search setting to focus on the quality of proof steps generated by the model itself.
As shown in Table \ref{tab:model_comparison}, vanilla base models such as Mistral-7B and Qwen3-1.7B struggle with formal proof tasks, yielding negligible success rates even when augmented with tree search.
This underscores the necessity of our domain-specific fine-tuning.

To evaluate the impact of the step revision mechanism, we analyze the theorems that become difficult to prove when the revision is disabled.
We observe that revision is most beneficial when the LLM fails to generate enough effective proof steps.
In particular, for the most challenging 77 theorems where the model initially proposed no valid steps and therefore failed immediately, the revision procedure is highly effective, enabling 24.7\% of these theorems to be proved.
For a broader set of 220 theorems where the LLM proposed fewer than 5\% valid steps, the revision mechanism still provides substantial gains, allowing 11.8\% of them to succeed.

Regarding the necessity of proof state filtering, particularly the counterexample detection, our experiments show that incorporating this component allows the system to identify an average of 1.3 counterexamples per evaluated theorem.
For the subset of search traces that contain counterexamples, the average number rises to 11.1.
These results indicate that counterexample detection is valuable for identifying unprovable states early and preventing unproductive search efforts, although it does not directly increase the overall success rate.


\begin{tcolorbox}[
  colback=gray!5,
  colframe=black!60,
  title=Response to RQ4
]
Each of the symbolic and neural components in our framework plays an essential role, and their synergy substantially strengthens the overall theorem-proving capability for software verification.
\end{tcolorbox}

\begin{table}[t]
\centering\small
\setlength{\tabcolsep}{2.5pt}
\caption{Proof success rate (\%) on several variants of our framework. 
The proof success rate increases progressively as tree search, retrieval augmentation, and hammer are incorporated, indicating the necessity of each component. 
The comparison highlights that our full approach, which replaces the off-the-shelf model with a fine-tuned LLM, achieves substantially stronger performance.}
\label{tab:api test}
\begin{tabular*}{\columnwidth}{l@{\extracolsep{\fill}}r}
\toprule
\textbf{Method Variation} & \textbf{Rate} \\
\midrule
DeepSeek (DpSk) v3.2 & 5.5 \\
DpSk + Tree Search & 12.5 \\
DpSk + RAG + Tree Search & 33.0 \\ 
DpSk + RAG + Tree Search + Hammer & 52.0 \\
\midrule
\textbf{Ours (Trained LLM \!+\! Tree Search \!+\! Hammer)} & \textbf{70.0} \\
\bottomrule
\end{tabular*}
\end{table}

\begin{table}[t]
\centering\small
\setlength{\tabcolsep}{2.5pt}
\caption{Effect of supervised fine-tuning under the same tree-search
configuration. Without fine-tuning, the base models solve almost no
theorems; in contrast, after supervised fine-tuning, both models reach around 60\%
proof success.}
\label{tab:model_comparison}
\begin{tabular*}{\columnwidth}{l@{\extracolsep{\fill}}r}
\toprule
\textbf{Method Variation} & \textbf{Rate} \\
\midrule
Mistral-7B + Tree Search  & 0.0 \\
Qwen3-1.7B + Tree Search  & 0.1 \\
\midrule
\textbf{Mistral-7B SFT + Tree Search}  & \textbf{59.8} \\
\textbf{Qwen3-1.7B SFT + Tree Search}  & \textbf{57.1} \\
\bottomrule
\end{tabular*}
\end{table}

\section{Further Analysis} \label{sect:analysis}

Beyond aggregate accuracy, we examine how our proposed framework actually
constructs successful proofs and what causes the remaining failures.

\subsection{Successful Case Studies} \label{sect:case-study}


We discuss two successful proofs that exhibit complementary patterns: a
short symmetry lemma closed by an LLM-proposed case analysis combined
with the symbolic tool Sledgehammer (C1), and a long Hoare-triple proof
for an seL4 fault handler built entirely from LLM-generated steps (C2).
For each, we show the theorem, the proof our framework discovered, and
the search behavior behind it.

\smallskip
\noindent\textbf{C1. A symmetry lemma in CSpace.}
The seL4 lemma \lstinline[style=isa]{same_object_as_commute}, from the
CSpace invariant development, states that the capability-equivalence
relation \lstinline[style=isa]{same_object_as} is symmetric. The seL4
proof reduces the equality to a one-way implication via
\lstinline[style=isa]{subgoal_tac} and closes both directions with a
single \lstinline[style=isa]{auto} call. Our framework takes a different
route: it splits on \lstinline[style=isa]{c'} first and lets Sledgehammer
discharge each constructor case independently.

\begin{isacode-green}
lemma same_object_as_commute:
  "same_object_as c' c = same_object_as c c'"
apply (case_tac c')
apply (simp_all add: same_object_as_def split: cap.splits)
apply blast
apply meson
apply blast
using bits_of_def apply force
apply blast
apply blast
using same_aobject_as_commute by blast
\end{isacode-green}

\emph{What the framework did.}
At depth~0, the LLM proposed several decompositions. 
Tree search explored seven of them, including \lstinline[style=isa]{case_tac c} with various
simp-set expansions, before settling on \lstinline[style=isa]{case_tac c'}
followed by \lstinline[style=isa]{simp_all (split: cap.splits)}, which
left a small frontier of constructor cases. Sledgehammer then closed the
leaves with single-line \lstinline[style=isa]{blast}, \lstinline[style=isa]{meson},
and \lstinline[style=isa]{force} calls, together with the previously
proved \lstinline[style=isa]{same_aobject_as_commute}.

\smallskip
\noindent\textbf{C2. Invariance under VM-fault handling.}
The lemma \lstinline[style=isa]{hv_inv_ex}, from the ARM
\lstinline[style=isa]{ArchSyscall_AI} session, is a top-level Hoare triple
for seL4's VM-fault handler: it asserts that an arbitrary precondition $P$
is preserved along the exceptional return path. The seL4 proof is a single
composite tactic that relies on dedicated \lstinline[style=isa]{*_inv}
weakest-precondition lemmas:

\begin{isacode-red}
lemma hv_inv_ex:
  "⦃P⦄ handle_vm_fault t vp ⦃λ_ _. True⦄, ⦃λ_. P⦄"
  by (cases vp;
      wpsimp wp: dmo_inv getDFSR_inv getFAR_inv
                 getIFSR_inv getRestartPC_inv)
\end{isacode-red}

Our framework arrives at a substantially different proof. It does
not use any of the \lstinline[style=isa]{*_inv} lemmas; instead, it
rebuilds the argument as a 29-step script of
\lstinline[style=isa]{wp}/\lstinline[style=isa]{wpsimp} calls,
\lstinline[style=isa]{clarsimp} unfoldings, and four explicit
postcondition weakenings via \lstinline[style=isa]{hoare_post_imp}.
Sledgehammer is never invoked: every step is produced by the LLM and
validated by the Isabelle kernel. The proof below is abridged.

\emph{What the framework did.}
After the initial \lstinline[style=isa]{cases vp} split, the proof unfolds
along three near-parallel branches, one per fault sub-handler
(\lstinline[style=isa]{getDFSR}, \lstinline[style=isa]{getFAR},
\lstinline[style=isa]{getIFSR}). Tree search repeatedly favored
\lstinline[style=isa]{wp}/\lstinline[style=isa]{wpsimp} steps interleaved
with \lstinline[style=isa]{clarsimp} unfoldings as the symbolic state
became exposed. At four points, the LLM proposed intermediate
postconditions through \lstinline[style=isa]{rule_tac P=...} +
\lstinline[style=isa]{hoare_post_imp}, which were essential to keep the
proof state tractable. None of the 29 steps required Sledgehammer.

\begin{isacode-green}
lemma hv_inv_ex:
  "⦃P⦄ handle_vm_fault t vp ⦃λ_ _. True⦄, ⦃λ_. P⦄"
apply (cases vp, simp_all)
apply wp
apply (clarsimp simp: do_machine_op_def getDFSR_def)
apply wp
apply (case_tac rv)
apply clarsimp
apply (rule_tac P="P and (λx. snd (aa,ba) = 
       machine_state x)" in hoare_post_imp)
(* ... 20 proof lines omitted ... *)
apply clarsimp
apply (drule in_inv_by_hoareD [OF gets_inv])
by simp
\end{isacode-green}

\smallskip
\noindent\textbf{Summary of successful cases.}
In both cases, the framework reaches the theorem through a proof
structure that differs from the seL4 script: decomposing the equality
on the opposite side in C1, and expanding a one-line composite tactic
into a 29-step derivation in C2. These examples show that the system
genuinely composes LLM generation, symbolic validation, and Sledgehammer
integration on non-trivial seL4 goals, rather than replaying memorized
scripts. Further examples are available in our GitHub
repository.

\subsection{Failure Analysis} \label{sect:failure-analysis}


We next turn to the failures. Across the three evaluation splits, 692
theorems remain unsolved (validation: 241, test: 199, test-hard: 252).
We study them at two different granularities. At the step level, our
\emph{tactic-error} analysis draws on the 7.77M individual tactic
attempts logged for the 241 failed theorems on the validation split,
where each rejected step records the underlying Isabelle kernel error.
At the run level, our \emph{search-outcome} analysis draws on the 451
failed runs from the test and test-hard splits, whose logs record the
maximum search depth reached before the budget was exhausted.

\smallskip
\noindent\textbf{Step-level error categories.}
Of the 7.77M tactic attempts on the validation failures, 30.1\% advance or
duplicate the proof state, while 69.8\% raise a kernel error.
Table~\ref{tab:err-cat} groups these errors into 12 categories, two of
which dominate.
The first is \emph{lemma hallucination}, which combines the
\lstinline[style=isa]{undefined-fact}, \lstinline[style=isa]{undefined-method},
and \lstinline[style=isa]{undefined-constant} categories and accounts for
61.1\% of all errors.
The second is \emph{shape mismatch}, captured by
\lstinline[style=isa]{tactic-inapplicable}, which contributes another
17.9\%.
Together, these two patterns explain 79\% of rejected tactics: the model
most often cites premises absent from the active context, and otherwise
produces well-formed tactics that do not fit the current goal shape.
The remaining categories, including schematic-variable mistakes, syntax
errors, type errors, tactic timeouts, and unfinished
\lstinline[style=isa]{by} closures, each contribute under 3\%.

\begin{table}[t]
\centering\footnotesize
\setlength{\tabcolsep}{2.5pt}
\caption{Step-level error breakdown over all tactic attempts on the
validation failures (241 theorems, 7.77M attempts). Undefined-fact alone accounts
for nearly two-thirds of all errors: the LLM repeatedly cites premises that
do not exist in the active context.}
\begin{tabular*}{\columnwidth}{l@{\extracolsep{\fill}}rrlrr}
\toprule
\textbf{Error category} & \textbf{Count} & \textbf{\%} &
\textbf{Error category} & \textbf{Count} & \textbf{\%} \\
\midrule
Undefined fact       & 3{,}263{,}494 & 60.1 & Bad method args.   & 65{,}430 & 1.2 \\
Tactic inapplicable  &   976{,}398 & 17.9 & Inner syntax error & 50{,}033 & 0.9 \\
Multi-line / other   &   423{,}982 &  7.8 & Undefined method   & 47{,}163 & 0.8 \\
Unknown variable     &   161{,}842 &  2.9 & Proof unfinished   & 37{,}100 & 0.6 \\
Outer syntax error   &   153{,}037 &  2.8 & THM exc.           & 26{,}750 & 0.4 \\
Type unification     &   110{,}717 &  2.0 & Undefined constant &  8{,}852 & 0.1 \\
Tactic timeout       &   102{,}099 &  1.8 &                    &          &     \\
\midrule
\multicolumn{4}{r}{All errors} & 5{,}427{,}419 & 100 \\
\bottomrule
\end{tabular*}

\label{tab:err-cat}
\end{table}

\smallskip
\noindent\textbf{Search-level outcomes.}
Table~\ref{tab:search-cat} shows how far search gets before exhausting its
budget. Only 3.7\% of failures stall by depth~10. Most failures therefore
make real multi-step progress ($\ge 11$ depth in 96\% of cases) but run
out of the 128-attempt budget later in the search. On the validation
split, where QuickCheck signals are logged, 23 of 241 failed theorems
(9.5\%) contain at least one search node with a genuine counterexample.
These cases suggest that the extracted obligation is unprovable in the
active context, typically because an essential assumption is dropped
when proof states are sliced from the original seL4 scripts to form the
benchmark; we treat them as benchmark-side issues rather than framework
failures. Sledgehammer was invoked on all 241 validation failures that
still had a residual goal but closed none of them, indicating that the
remaining goals lie outside the reach of off-the-shelf SMT/superposition.

\begin{table}[t]
\centering\small
\setlength{\tabcolsep}{2.5pt}
\caption{Search-depth distribution for failures on the test and
test-hard splits (the validation log format does not record
depth markers). Only 3.7\% of failures stall before depth~10; most
failures make substantial multi-step progress and run out of budget
later in the search.}
\begin{tabular*}{\columnwidth}{l@{\extracolsep{\fill}}rr}
\toprule
\textbf{Max search depth reached} & \textbf{Theorems} & \textbf{\%} \\
\midrule
$0$ (no progress past initial state)        &  11 &  2.4 \\
$1$--$10$ (early stall)                      &   6 &  1.3 \\
$11$--$30$ (medium-depth stall)              & 109 & 24.2 \\
$31$--$60$ (deep stall)                      & 181 & 40.1 \\
$> 60$ (very deep)                           & 145 & 32.0 \\
\midrule
Total                                        & 451 & 100 \\
\bottomrule
\end{tabular*}
\label{tab:search-cat}
\end{table}

\noindent\textbf{Summary of failure modes.}
The two analyses converge on the same bottlenecks. At the step level,
wasted budget is dominated by missing premises ($\sim$61\% of errors)
and goal-shape mismatches ($\sim$18\%). At the run level, most failed
runs do not stall early; they reach substantial depth and then exhaust
the budget on residual goals that Sledgehammer cannot close either.
We therefore see two clear directions for future work: better premise
grounding to reduce lemma hallucination, and longer-horizon planning to
prevent late-search collapse on deep proofs.

\section{Related Work} \label{sect:related}

\subsection{Software System Verification via ITP}

Formal methods have successfully verified critical software systems 
including operating system kernels, hypervisors, file systems and compilers. 
Interactive theorem proving (ITP) provides strong formal assurance by allowing engineers to encode specifications and build proofs in ITP systems such as Rocq or Isabelle/HOL. 
Notable examples include CertiKOS~\citep{gu2016certikos}, which employs a compositional refinement framework to verify a layered OS kernel; CompCert~\citep{leroy2016compcert}, an optimizing C compiler whose translation passes are all verified to preserve semantics;  FSCQ~\citep{fscq}, a crash-safe file system proven correct under all failure scenarios using Crash Hoare Logic; seKVM hypervisor~\citep{sekvm}, 
a reduced KVM core with enforced confidentiality and integrity of guest memory; seL4~\citep{klein2014comprehensive}, an OS microkernel with machine-checked proofs of functional correctness and security properties.
These efforts demonstrate that ITP enables high-assurance proofs for critical systems, albeit at a significant engineering cost.

\subsection{Automated Theorem Proving}

Automated theorem proving (ATP) has long served as a benchmark for machine reasoning. Classical ATP systems rely on symbolic methods, such as resolution-based first-order provers (e.g., Vampire~\citep{riazanov2002design}, E~\citep{schulz2019faster}) and SMT solvers (e.g., Z3~\citep{de2008z3}, CVC5~\citep{barbosa2022cvc5}) with built-in theories. These solvers are also integrated into interactive theorem provers via hammers (e.g., Sledgehammer~\citep{bohme2010sledgehammer} in Isabelle), enabling automated discharge of proof goals~\citep{czajka2018hammer}.
Related to ATP, several works (e.g., Ironclad Apps~\citep{hawblitzel2014ironclad} and Verve~\citep{yang2011safe}) leverage verification-aware languages (e.g., Dafny~\citep{leino2010dafny} and Boogie~\citep{barnett2005boogie}) to prove memory and control-flow safety of system components, including kernels and device drivers. 

In recent years, machine learning has been increasingly applied to
ATP~\citep{li2024survey}. Early efforts focused on premise selection and
tactic prediction using classical learning methods (e.g.,
Holophrasm~\citep{whalen2016holophrasm},
TacticToe~\citep{gauthier2021tactictoe},
Proverbot9001~\citep{sanchez2020proverbot},
Diva~\citep{first2022diversity}), showing improved success rates in
Metamath, HOL4, and Coq. More recently, large language models (LLMs)
have been used to generate complete proofs or
sketches~\citep{jiang2022draft,cao2025reviving}, achieving near-superhuman
results on mathematical problems~\citep{trinh2024solving}. Notably,
AlphaProof~\citep{hubert2025olympiad} combines Monte Carlo tree search
with reinforcement learning to solve IMO-level competition problems in
Lean, representing the state of the art in search-based mathematical
theorem proving.

Closer to our setting, prior efforts have attempted to bring such
techniques to software verification, including Baldur~\citep{first2023baldur},
PALM~\citep{lu2024proof}, Guided Proof
Search~\citep{prasad2024guided}, and
Goedel-Code-Prover~\citep{li2026goedel}. However, these approaches
remain limited to textbook-scale problems and do not address the
challenges of real-world system-level verification projects such as
seL4.

Complementary to whole-proof generation, the proof-step prediction paradigm harnesses LLMs to propose the next tactic or lemma, guided by proof-state feedback. Systems such as GPT-f~\citep{polu2020generative}, LISA~\citep{jiang2021lisa}, HTPS~\citep{lample2022hypertree} and LeanDojo~\citep{yang2023leandojo} demonstrate the potential of this paradigm across Metamath, Isabelle and Lean, but again primarily on isolated mathematical problems. Our method fills this gap by integrating a fine-tuned LLM for proof search with symbolic techniques, namely proof-step revision and proof-state filtering, making large-scale verification both effective and efficient.

To further enhance reliability, hybrid approaches are proposed to incorporate symbolic provers into LLM pipelines. 
A series of works, including Thor~\citep{jiang2022thor}, ProofAug~\citep{liu2025proofaug}, Strat2Rocq~\citep{fang2025proof}, and LIPS~\citep{li2025proving}, exemplify this by coupling LLM-generated reasoning with ATPs or hammers. 
However, hammer-only proofs tend to grow brittle on long, multi-step
obligations and place additional trust in the external solver. In
contrast, our framework integrates LLMs with an ITP-driven proof search
and invokes ATP only as a final-stage closer for residual subgoals,
keeping the bulk of the proof inside the kernel-checked workflow.

\section{Limitations and Future Work} \label{sect:lim}

Despite strong empirical performance, our framework has several limitations that suggest directions for future research.

\noindent{\em Computational cost}. The overall tree-search procedure remains computationally expensive. Each iteration requires invoking Isabelle for proof-step execution and proof-state checking, which 
results in substantial runtime overhead. 
Future work includes developing efficient state caching, parallel expansion, and cost-aware search policies to minimize unnecessary exploration and improve throughput.

\noindent{\em Performance degradation for longer proofs}. While our method successfully constructs longer proofs than prior methods, its performance indeed degrades as proof length increases. The current tree-search procedure adopts one-step prediction and bounded search, making it challenging to uncover deeper reasoning chains. Future improvements could involve 
hierarchical planning, subgoal decomposition, or reinforcement learning from bootstrapped trajectories. 

\noindent{\em Rigid rule-driven step revision}. The proof step revision modules heavily rely on heuristic tactic-premise recombination and lightweight premise retrieval. 
These mechanisms, though helpful and portable via statistical extraction, are imperfect and may 
introduce misleading candidates. 
Future improvements may include learning the relevance of premises directly from the proof state or integrating stronger semantic filters.

\noindent{\em Leveraging frontier models}. As frontier LLMs continue to advance in reasoning capabilities, an important direction is to explore whether agentic workflows, where a large model autonomously invokes verification tools, inspects proof states, and iterates, can complement or even replace parts of our pipeline. Our current use of fine-tuned small models is motivated by two practical advantages: lower cost and higher inference throughput, both critical for generating and scoring large candidate pools during tree search. A promising hybrid architecture would delegate high-level proof planning and subgoal decomposition to a frontier model, directly addressing the performance degradation on longer proofs discussed above, while relying on lightweight fine-tuned models for intensive step generation and ranking.


\section{Conclusion} \label{sect:conc}
In this paper, we present a neuro-symbolic proof-generation framework that automates proof search for large-scale software verification projects in interactive theorem provers. Instead of attempting whole-proof synthesis, our approach performs a best-first tree search over fine-grained proof states, repeatedly querying a lightweight, fine-tuned LLM for the next proof step. This paradigm not only substantially improves data efficiency, but also enables a seamless integration of symbolic reasoning tools  
with LLM-based step generation and scoring, forming an effective and efficient pipeline for proof-step construction as well as proof-state filtering and ranking. 
%
The experiments on the seL4 verification corpus show that our approach 
significantly outperforms prior 
baselines. 
Notably, our method succeeds on a considerably higher proportion of multi-step, nontrivial theorems, addressing a long-standing limitation of LLM-based provers. Additional experiments on AFP projects and translated program-verification benchmarks further showcase strong cross-project generalizability. 

Our work provides compelling evidence that proof-step–centric neuro-symbolic search can serve as a practical path toward scalable, automated software verification at the systems level, e.g., operating systems kernels, network architecture, and databases. While significant challenges remain, our results represent promising progress toward developing AI-enabled, rigorous engineering methods for real-world computer systems.



\medskip
\noindent\textbf{Availability.}
The tool, experimental data, and source code are available at 
\href{https://github.com/SoaringE/seL4-proof-search}{the seL4-proof-search GitHub repository}. 
Additionally, a prebuilt Docker image containing the seL4 verification environment and all required dependencies is available at 
\href{https://hub.docker.com/r/soaringe/sel4-test}{Docker Hub}.


\smallskip
\noindent\textbf{Acknowledgments.}
We thank the anonymous reviewers and our shepherd for their valuable and constructive feedback.
This work is supported by the Fundamental and Interdisciplinary Disciplines Breakthrough Plan of the Ministry of Education of China (No.\ JYB2025XDXM118), the National Natural Science Foundation of China (Grant \#62025202), and the Frontier Technologies R\&D Program of Jiangsu (BF2024059).  T. Chen is partially supported by overseas grants from the State Key
Laboratory of Novel Software Technology, Nanjing University (KFKT2023A04, KFKT2025A05).

\clearpage
\newpage 

\balance
\bibliographystyle{unsrt}
\bibliography{references}

@article{li2026goedel,
  title={Goedel-Code-Prover: Hierarchical Proof Search for Open State-of-the-Art Code Verification},
  author={Li, Zenan and Yang, Ziran and Zhao, Haoyu and Zhao, Andrew and Tang, Shange and Yang, Kaiyu and Gupta, Aarti and Su, Zhendong and Jin, Chi and others},
  journal={arXiv preprint arXiv:2603.19329},
  year={2026}
}

@inproceedings{li2025proving,
  title={Proving Olympiad Inequalities by Synergizing LLMs and Symbolic Reasoning},
  author={Li, Zenan and Li, Zhaoyu and Tang, Wen and Zhang, Xian and Yao, Yuan and Si, Xujie and Yang, Fan and Yang, Kaiyu and Ma, Xiaoxing},
  booktitle={The Thirteenth International Conference on Learning Representations},
  year={2025}
}

@article{hubert2025olympiad,
  title={Olympiad-level formal mathematical reasoning with reinforcement learning},
  author={Hubert, Thomas and Mehta, Rishi and Sartran, Laurent and Horv{\'a}th, Mikl{\'o}s Z and {\v{Z}}u{\v{z}}i{\'c}, Goran and Wieser, Eric and Huang, Aja and Schrittwieser, Julian and Schroecker, Yannick and Masoom, Hussain and others},
  journal={Nature},
  pages={1--3},
  year={2025},
  publisher={Nature Publishing Group UK London}
}

@article{blanchette2016hammering,
  title={Hammering towards QED},
  author={Blanchette, Jasmin Christian and Kaliszyk, Cezary and Paulson, Lawrence C and Urban, Josef},
  journal={Journal of Formalized Reasoning},
  volume={9},
  number={1},
  pages={101--148},
  year={2016}
}

@inproceedings{barnett2005boogie,
  title={Boogie: A modular reusable verifier for object-oriented programs},
  author={Barnett, Mike and Chang, Bor-Yuh Evan and DeLine, Robert and Jacobs, Bart and Leino, K Rustan M},
  booktitle={International Symposium on Formal Methods for Components and Objects},
  pages={364--387},
  year={2005},
  organization={Springer}
}

@inproceedings{leino2010dafny,
  title={Dafny: An automatic program verifier for functional correctness},
  author={Leino, K Rustan M},
  booktitle={International conference on logic for programming artificial intelligence and reasoning},
  pages={348--370},
  year={2010},
  organization={Springer}
}

@article{yang2011safe,
  title={Safe to the last instruction: automated verification of a type-safe operating system},
  author={Yang, Jean and Hawblitzel, Chris},
  journal={Communications of the ACM},
  volume={54},
  number={12},
  pages={123--131},
  year={2011},
  publisher={ACM New York, NY, USA}
}

@inproceedings{hawblitzel2014ironclad,
  title={Ironclad apps:$\{$End-to-End$\}$ security via automated $\{$Full-System$\}$ verification},
  author={Hawblitzel, Chris and Howell, Jon and Lorch, Jacob R and Narayan, Arjun and Parno, Bryan and Zhang, Danfeng and Zill, Brian},
  booktitle={11th USENIX symposium on operating systems design and implementation (OSDI 14)},
  pages={165--181},
  year={2014}
}

@inproceedings{gu2016certikos,
  title={$\{$CertiKOS$\}$: An extensible architecture for building certified concurrent $\{$OS$\}$ kernels},
  author={Gu, Ronghui and Shao, Zhong and Chen, Hao and Wu, Xiongnan Newman and Kim, Jieung and Sj{\"o}berg, Vilhelm and Costanzo, David},
  booktitle={12th USENIX Symposium on Operating Systems Design and Implementation (OSDI 16)},
  pages={653--669},
  year={2016}
}

@article{robertson2009probabilistic,
  title={The probabilistic relevance framework: BM25 and beyond},
  author={Robertson, Stephen and Zaragoza, Hugo and others},
  journal={Foundations and Trends{\textregistered} in Information Retrieval},
  volume={3},
  number={4},
  pages={333--389},
  year={2009},
  publisher={Now Publishers, Inc.}
}

@article{jaccard1901etude,
  title={{\'E}tude comparative de la distribution florale dans une portion des Alpes et des Jura},
  author={Jaccard, Paul},
  journal={Bull Soc Vaudoise Sci Nat},
  volume={37},
  pages={547--579},
  year={1901}
}

@inproceedings{levenshtein1966binary,
  title={Binary codes capable of correcting deletions, insertions, and reversals},
  author={Levenshtein, Vladimir I and others},
  booktitle={Soviet physics doklady},
  volume={10},
  number={8},
  pages={707--710},
  year={1966},
  organization={Soviet Union}
}

@inproceedings{wu2024internlm2,
  title={InternLM2. 5-StepProver: Advancing Automated Theorem Proving via Critic-Guided Search},
  author={Wu, Zijian and Huang, Suozhi and Zhou, Zhejian and Ying, Huaiyuan and Yuan, Zheng and Zhang, Wenwei and Lin, Dahua and Chen, Kai},
  booktitle={2nd AI for Math Workshop@ ICML 2025},
  year={2025}
}

@article{Clarke1996,
  author    = {Edmund M. Clarke and Jeannette M. Wing},
  title     = {Formal methods: state of the art and future directions},
  journal   = {ACM Computing Surveys},
  volume    = {28},
  number    = {4},
  pages     = {626--643},
  year      = {1996},
  doi       = {10.1145/242223.242257}
}

@article{Woodcock2009,
  author    = {James Woodcock and Peter Gorm Larsen and Juan Bicarregui and John Fitzgerald},
  title     = {Formal Methods: Practice and Experience},
  journal   = {ACM Computing Surveys},
  volume    = {41},
  number    = {4},
  pages     = {1--36},
  year      = {2009},
  doi       = {10.1145/1592434.1592436}
}

@inproceedings{Blanchet2003,
  author    = {Bruno Blanchet and Patrick Cousot and Radhia Cousot and J{\'e}r{\^o}me Feret and Laurent Mauborgne and Antoine Min{\'e} and David Monniaux and Xavier Rival},
  title     = {A static analyzer for large safety-critical software},
  booktitle = {Proceedings of the ACM SIGPLAN Conference on Programming Language Design and Implementation (PLDI)},
  year      = {2003},
  pages     = {196--207},
  doi       = {10.1145/780822.781153}
}

@article{dubey2024llama,
  title={The llama 3 herd of models},
  author={Dubey, Abhimanyu and Jauhri, Abhinav and Pandey, Abhinav and Kadian, Abhishek and Al-Dahle, Ahmad and Letman, Aiesha and Mathur, Akhil and Schelten, Alan and Yang, Amy and Fan, Angela and others},
  journal={arXiv e-prints},
  pages={arXiv--2407},
  year={2024}
}

@article{jiang2023mistral,
  title   = {Mistral 7B},
  author  = {Albert Q. Jiang and Alexandre Sablayrolles and Arthur Mensch and Chris Bamford and Devendra Singh Chaplot and Diego de las Casas and Florian Bressand and Gianna Lengyel and Guillaume Lample and Lucile Saulnier and Lélio Renard Lavaud and Marie-Anne Lachaux and Pierre Stock and Teven Le Scao and Thibaut Lavril and Thomas Wang and Timothée Lacroix and William El Sayed},
  journal = {arXiv preprint arXiv:2310.06825},
  year    = {2023},
  doi     = {10.48550/arXiv.2310.06825},
  url     = {https://arxiv.org/abs/2310.06825}
}

@article{yang2025qwen3,
  title={Qwen3 technical report},
  author={Yang, An and Li, Anfeng and Yang, Baosong and Zhang, Beichen and Hui, Binyuan and Zheng, Bo and Yu, Bowen and Gao, Chang and Huang, Chengen and Lv, Chenxu and others},
  journal={arXiv preprint arXiv:2505.09388},
  year={2025}
}

@misc{openai2025gpt51,
  author = {OpenAI},
  title  = {{GPT-5.1}: A Smarter, More Conversational {ChatGPT}},
  year   = {2025},
  url    = {https://openai.com/index/gpt-5-1/},
  note   = {Accessed model variants: GPT-5.1 Instant and GPT-5.1 Thinking}
}

@misc{google2024gemini,
  title        = {Gemini: A family of highly capable multimodal models},
  author       = {{Google DeepMind}},
  year         = {2024},
  howpublished = {\url{https://deepmind.google/technologies/gemini/}},
  note         = {Accessed model variant: Gemini 3 Pro}
}

@inproceedings{zheng2024llamafactory,
  title={LlamaFactory: Unified Efficient Fine-Tuning of 100+ Language Models},
  author={Yaowei Zheng and Richong Zhang and Junhao Zhang and Yanhan Ye and Zheyan Luo and Zhangchi Feng and Yongqiang Ma},
  booktitle={Proceedings of the 62nd Annual Meeting of the Association for Computational Linguistics (Volume 3: System Demonstrations)},
  address={Bangkok, Thailand},
  publisher={Association for Computational Linguistics},
  year={2024},
  url={http://arxiv.org/abs/2403.13372}
}

@article{nishant2023py4j,
  author    = {N. M. Nishant and G. S. Mamatha},
  title     = {Using Py4J for Java-Python Communication},
  journal   = {International Journal of Advanced Research in Computer and Communication Engineering (IJARCCE)},
  year      = {2023},
  doi       = {10.17148/IJARCCE.2023.12610}
}

@misc{unruh2025scalaisabelle,
  author       = {Dominique Unruh},
  title        = {scala-isabelle: A Scala library for interacting with Isabelle},
  year         = {2025},
  howpublished = {\url{https://github.com/dominique-unruh/scala-isabelle}},
  note         = {Accessed 2025-11-18}
}

@misc{SolveDirect2025,
  author       = {Timothy Bourke and Gerwin Klein},
  title        = {Solve\_Direct: A tool for Isabelle/HOL to check whether a newly stated theorem can be solved directly by an existing theorem},
  year         = {2025},
  howpublished = {\url{https://isabelle.in.tum.de/library/HOL/HOL/ISABELLE\_HOME/src/Tools/solve\_direct.ML.html}},
  note         = {File: src/Tools/solve\_direct.ML in Isabelle/HOL library},
}

@inproceedings{piotrowski2023machine,
  title={Machine-learned premise selection for Lean},
  author={Piotrowski, Bartosz and Mir, Ramon Fern{\'a}ndez and Ayers, Edward},
  booktitle={International Conference on Automated Reasoning with Analytic Tableaux and Related Methods},
  pages={175--186},
  year={2023},
  organization={Springer}
}

@article{wu2020neural,
  title={Neural theorem proving on inequality problems},
  author={Wu, Yuhuai and Jiang, Albert and Grosse, Roger and Ba, Jimmy},
  journal={Artificial Intelligence and Theorem Proving (AITP)},
  year={2020}
}

@inproceedings{huang2025leanprogress,
  title={LeanProgress: Guiding Search for Neural Theorem Proving via Proof Progress Prediction},
  author={Huang, Suozhi and Song, Peiyang and George, Robert Joseph and Anandkumar, Anima},
  booktitle={ICLR 2025 Workshop: VerifAI: AI Verification in the Wild},
  year={2025}
}

@article{wu2021tacticzero,
  title={Tacticzero: Learning to prove theorems from scratch with deep reinforcement learning},
  author={Wu, Minchao and Norrish, Michael and Walder, Christian and Dezfouli, Amir},
  journal={Advances in Neural Information Processing Systems},
  volume={34},
  pages={9330--9342},
  year={2021}
}

@inproceedings{prasad2024guided,
  title={Guided Proof Search Using Large Language Models and Lemma Extraction in Coq},
  author={Prasad, Tarun and Amin, Nada},
  booktitle={ICLR 2025 Workshop: VerifAI: AI Verification in the Wild},
  year={2024}
}

@article{trinh2024solving,
  title={Solving olympiad geometry without human demonstrations},
  author={Trinh, Trieu H and Wu, Yuhuai and Le, Quoc V and He, He and Luong, Thang},
  journal={Nature},
  volume={625},
  number={7995},
  pages={476--482},
  year={2024},
  publisher={Nature Publishing Group UK London}
}

@inproceedings{cao2025reviving,
  title={Reviving DSP for Advanced Theorem Proving in the Era of Reasoning Models},
  author={Cao, Chenrui and Song, Liangcheng and Li, Zenan and Le, Xinyi and Zhang, Xian and XUE, HUI and Yang, Fan},
  booktitle={The Thirty-ninth Annual Conference on Neural Information Processing Systems},
  year={2025}
}

@inproceedings{li2024survey,
  title={A Survey on Deep Learning for Theorem Proving},
  author={Li, Zhaoyu and Sun, Jialiang and Murphy, Logan and Su, Qidong and Li, Zenan and Zhang, Xian and Yang, Kaiyu and Si, Xujie},
  booktitle={Proceedings of the First Conference on Language Modeling},
  year={2024},
}

@inproceedings{barbosa2022cvc5,
  title={cvc5: A versatile and industrial-strength SMT solver},
  author={Barbosa, Haniel and Barrett, Clark and Brain, Martin and Kremer, Gereon and Lachnitt, Hanna and Mann, Makai and Mohamed, Abdalrhman and Mohamed, Mudathir and Niemetz, Aina and N{\"o}tzli, Andres and others},
  booktitle={International Conference on Tools and Algorithms for the Construction and Analysis of Systems},
  pages={415--442},
  year={2022},
  organization={Springer}
}

@inproceedings{claessen2000quickcheck,
  title={QuickCheck: a lightweight tool for random testing of Haskell programs},
  author={Claessen, Koen and Hughes, John},
  booktitle={Proceedings of the fifth ACM SIGPLAN international conference on Functional programming},
  pages={268--279},
  year={2000}
}

@article{ringer2019qed,
  title={QED at large: A survey of engineering of formally verified software},
  author={Ringer, Talia and Palmskog, Karl and Sergey, Ilya and Gligoric, Milos and Tatlock, Zachary and others},
  journal={Foundations and Trends{\textregistered} in Programming Languages},
  volume={5},
  number={2-3},
  pages={102--281},
  year={2019},
  publisher={Now Publishers, Inc.}
}

@article{fang2025proof,
  title={Proof Strategy Extraction from LLMs for Enhancing Symbolic Provers},
  author={Fang, Jian and Sun, Yican and Xiong, Yingfei},
  journal={arXiv preprint arXiv:2510.10131},
  year={2025}
}

@article{riazanov2002design,
  title={The design and implementation of VAMPIRE},
  author={Riazanov, Alexandre and Voronkov, Andrei},
  journal={AI communications},
  volume={15},
  number={2-3},
  pages={91--110},
  year={2002},
  publisher={SAGE Publications Sage UK: London, England}
}

@inproceedings{schulz2019faster,
  title={Faster, higher, stronger: E 2.3},
  author={Schulz, Stephan and Cruanes, Simon and Vukmirovi{\'c}, Petar},
  booktitle={International Conference on Automated Deduction},
  pages={495--507},
  year={2019},
  organization={Springer}
}

@inproceedings{de2008z3,
  title={Z3: An efficient SMT solver},
  author={De Moura, Leonardo and Bj{\o}rner, Nikolaj},
  booktitle={International conference on Tools and Algorithms for the Construction and Analysis of Systems},
  pages={337--340},
  year={2008},
  organization={Springer}
}

@article{czajka2018hammer,
  title={Hammer for Coq: Automation for dependent type theory},
  author={Czajka, {\L}ukasz and Kaliszyk, Cezary},
  journal={Journal of automated reasoning},
  volume={61},
  number={1},
  pages={423--453},
  year={2018},
  publisher={Springer}
}

@inproceedings{bohme2010sledgehammer,
  title={Sledgehammer: judgement day},
  author={B{\"o}hme, Sascha and Nipkow, Tobias},
  booktitle={International Joint Conference on Automated Reasoning},
  pages={107--121},
  year={2010},
  organization={Springer}
}

@article{whalen2016holophrasm,
  title={Holophrasm: a neural automated theorem prover for higher-order logic},
  author={Whalen, Daniel},
  journal={arXiv preprint arXiv:1608.02644},
  year={2016}
}

@article{gauthier2021tactictoe,
  title={TacticToe: learning to prove with tactics},
  author={Gauthier, Thibault and Kaliszyk, Cezary and Urban, Josef and Kumar, Ramana and Norrish, Michael},
  journal={Journal of Automated Reasoning},
  volume={65},
  number={2},
  pages={257--286},
  year={2021},
  publisher={Springer}
}

@inproceedings{sanchez2020proverbot,
  title={Generating correctness proofs with neural networks},
  author={Sanchez-Stern, Alex and Alhessi, Yousef and Saul, Lawrence and Lerner, Sorin},
  booktitle={Proceedings of the 4th ACM SIGPLAN International Workshop on Machine Learning and Programming Languages},
  pages={1--10},
  year={2020}
}

@inproceedings{first2022diversity,
  title={Diversity-driven automated formal verification},
  author={First, Emily and Brun, Yuriy},
  booktitle={Proceedings of the 44th International Conference on Software Engineering},
  pages={749--761},
  year={2022}
}

@article{polu2020generative,
  title={Generative language modeling for automated theorem proving},
  author={Polu, Stanislas and Sutskever, Ilya},
  journal={arXiv preprint arXiv:2009.03393},
  year={2020}
}

@article{lample2022hypertree,
  title={Hypertree proof search for neural theorem proving},
  author={Lample, Guillaume and Lacroix, Timothee and Lachaux, Marie-Anne and Rodriguez, Aurelien and Hayat, Amaury and Lavril, Thibaut and Ebner, Gabriel and Martinet, Xavier},
  journal={Advances in neural information processing systems},
  volume={35},
  pages={26337--26349},
  year={2022}
}

@inproceedings{jiang2021lisa,
  title={Lisa: Language models of isabelle proofs},
  author={Jiang, Albert Qiaochu and Li, Wenda and Han, Jesse Michael and Wu, Yuhuai},
  booktitle={6th Conference on Artificial Intelligence and Theorem Proving},
  pages={378--392},
  year={2021}
}

@inproceedings{first2023baldur,
  title={Baldur: Whole-proof generation and repair with large language models},
  author={First, Emily and Rabe, Markus N and Ringer, Talia and Brun, Yuriy},
  booktitle={Proceedings of the 31st ACM Joint European Software Engineering Conference and Symposium on the Foundations of Software Engineering},
  pages={1229--1241},
  year={2023}
}

@article{yang2023leandojo,
  title={Leandojo: Theorem proving with retrieval-augmented language models},
  author={Yang, Kaiyu and Swope, Aidan and Gu, Alex and Chalamala, Rahul and Song, Peiyang and Yu, Shixing and Godil, Saad and Prenger, Ryan J and Anandkumar, Animashree},
  journal={Advances in Neural Information Processing Systems},
  volume={36},
  pages={21573--21612},
  year={2023}
}

@inproceedings{lu2024proof,
  title={Proof automation with large language models},
  author={Lu, Minghai and Delaware, Benjamin and Zhang, Tianyi},
  booktitle={Proceedings of the 39th IEEE/ACM International Conference on Automated Software Engineering},
  pages={1509--1520},
  year={2024}
}

@article{jiang2022thor,
  title={Thor: Wielding hammers to integrate language models and automated theorem provers},
  author={Jiang, Albert Qiaochu and Li, Wenda and Tworkowski, Szymon and Czechowski, Konrad and Odrzyg{\'o}{\'z}d{\'z}, Tomasz and Mi{\l}o{\'s}, Piotr and Wu, Yuhuai and Jamnik, Mateja},
  journal={Advances in Neural Information Processing Systems},
  volume={35},
  pages={8360--8373},
  year={2022}
}

@inproceedings{jiang2022draft,
  title={Draft, Sketch, and Prove: Guiding Formal Theorem Provers with Informal Proofs},
  author={Jiang, Albert Q and Welleck, Sean and Zhou, Jin Peng and Lacroix, Timothee and Liu, Jiacheng and Li, Wenda and Jamnik, Mateja and Lample, Guillaume and Wu, Yuhuai},
  booktitle={The Eleventh International Conference on Learning Representations (ICLR)},
  year={2023}
}

@inproceedings{leroy2016compcert,
  title={CompCert-a formally verified optimizing compiler},
  author={Leroy, Xavier and Blazy, Sandrine and K{\"a}stner, Daniel and Schommer, Bernhard and Pister, Markus and Ferdinand, Christian},
  booktitle={ERTS 2016: Embedded Real Time Software and Systems, 8th European Congress},
  year={2016}
}

@article{klein2014comprehensive,
  title={Comprehensive formal verification of an OS microkernel},
  author={Klein, Gerwin and Andronick, June and Elphinstone, Kevin and Murray, Toby and Sewell, Thomas and Kolanski, Rafal and Heiser, Gernot},
  journal={ACM Transactions on Computer Systems (TOCS)},
  volume={32},
  number={1},
  pages={1--70},
  year={2014},
  publisher={ACM New York, NY, USA}
}

@inproceedings{fscq,
  title={Using Crash Hoare logic for certifying the FSCQ file system},
  author={Chen, Haogang and Ziegler, Daniel and Chajed, Tej and Chlipala, Adam and Kaashoek, M Frans and Zeldovich, Nickolai},
  booktitle={Proceedings of the 25th Symposium on Operating Systems Principles},
  pages={18--37},
  year={2015}
}

@inproceedings{qin2025can,
  title={Can Large Language Models Verify System Software? A Case Study Using FSCQ as a Benchmark},
  author={Qin, Jianxing and Du, Alexander and Zhang, Danfeng and Lentz, Matthew and Zhuo, Danyang},
  booktitle={Proceedings of the 2025 Workshop on Hot Topics in Operating Systems},
  pages={34--41},
  year={2025}
}

@inproceedings{thompson2024rango,
  title={Rango: Adaptive Retrieval-Augmented Proving for Automated Software Verification},
  author={Thompson, Kyle and Saavedra, Nuno and Carrott, Pedro and Fisher, Kevin and Sanchez-Stern, Alex and Brun, Yuriy and Ferreira, Jo{\~a}o F and Lerner, Sorin and First, Emily},
  booktitle={2025 IEEE/ACM 47th International Conference on Software Engineering (ICSE)},
  pages={347--359},
  year={2025}
}

@inproceedings{zhang2024selene,
  title={Selene: Pioneering Automated Proof in Software Verification},
  author={Zhang, Lichen and Lu, Shuai and Duan, Nan},
  booktitle={Proceedings of the 62nd Annual Meeting of the Association for Computational Linguistics (ACL)},
  pages={1776--1789},
  year={2024}
}

@article{lin2024fvel,
  title={FVEL: Interactive formal verification environment with large language models via theorem proving},
  author={Lin, Xiaohan and Cao, Qingxing and Huang, Yinya and Wang, Haiming and Lu, Jianqiao and Liu, Zhengying and Song, Linqi and Liang, Xiaodan},
  journal={Advances in Neural Information Processing Systems},
  volume={37},
  pages={54932--54946},
  year={2024}
}

@inproceedings{matichuk2015cost,
author = {Matichuk, Daniel and Murray, Toby and Andronick, June and Jeffery, Ross and Klein, Gerwin and Staples, Mark},
title = {Empirical study towards a leading indicator for cost of formal software verification},
year = {2015},
isbn = {9781479919345},
publisher = {IEEE Press},
booktitle = {Proceedings of the 37th International Conference on Software Engineering - Volume 1},
pages = {722–732},
numpages = {11},
location = {Florence, Italy},
series = {ICSE '15}
}

@inproceedings{xin2025bfs,
  title={Bfs-prover: Scalable best-first tree search for llm-based automatic theorem proving},
  author={Xin, Ran and Xi, Chenguang and Yang, Jie and Chen, Feng and Wu, Hang and Xiao, Xia and Sun, Yifan and Zheng, Shen and Ding, Ming},
  booktitle={Proceedings of the 63rd Annual Meeting of the Association for Computational Linguistics (Volume 1: Long Papers)},
  pages={32588--32599},
  year={2025}
}

@inproceedings{kuhlwein2013mash,
  title={MaSh: machine learning for Sledgehammer},
  author={K{\"u}hlwein, Daniel and Blanchette, Jasmin Christian and Kaliszyk, Cezary and Urban, Josef},
  booktitle={International Conference on Interactive Theorem Proving},
  pages={35--50},
  year={2013},
  organization={Springer}
}

@article{meng2009lightweight,
  title={Lightweight relevance filtering for machine-generated resolution problems},
  author={Meng, Jia and Paulson, Lawrence C},
  journal={Journal of Applied Logic},
  volume={7},
  number={1},
  pages={41--57},
  year={2009},
  publisher={Elsevier}
}

@inproceedings{liu2025proofaug,
  title={ProofAug: Efficient Neural Theorem Proving via Fine-grained Proof Structure Analysis},
  author={Liu, Haoxiong and Sun, Jiacheng and Li, Zhenguo and Yao, Andrew C},
  booktitle={International Conference on Machine Learning},
  pages={39568--39586},
  year={2025},
  organization={PMLR}
}

@inproceedings{blanchette2010nitpick,
  title={Nitpick: A counterexample generator for higher-order logic based on a relational model finder},
  author={Blanchette, Jasmin Christian and Nipkow, Tobias},
  booktitle={International conference on interactive theorem proving},
  pages={131--146},
  year={2010},
  organization={Springer}
}

@article{X86_Semantics-AFP,
  author  = {Freek Verbeek and Abhijith Bharadwaj and Joshua Bockenek and Ian Roessle and Timmy Weerwag and Binoy Ravindran},
  title   = {X86 instruction semantics and basic block symbolic execution},
  journal = {Archive of Formal Proofs},
  month   = {October},
  year    = {2021},
  note    = {\url{https://isa-afp.org/entries/X86_Semantics.html},
             Formal proof development},
  ISSN    = {2150-914x},
}

@article{IEEE_Floating_Point-AFP,
  author  = {Lei Yu},
  title   = {A Formal Model of IEEE Floating Point Arithmetic},
  journal = {Archive of Formal Proofs},
  month   = {July},
  year    = {2013},
  note    = {\url{https://isa-afp.org/entries/IEEE_Floating_Point.html},
             Formal proof development},
  ISSN    = {2150-914x},
}

@article{SATSolverVerification-AFP,
  author  = {Filip Marić},
  title   = {Formal Verification of Modern SAT Solvers},
  journal = {Archive of Formal Proofs},
  month   = {July},
  year    = {2008},
  note    = {\url{https://isa-afp.org/entries/SATSolverVerification.html},
             Formal proof development},
  ISSN    = {2150-914x},
}

@inproceedings {sekvm,
  author = {Shih-Wei Li and Xupeng Li and Ronghui Gu and Jason Nieh and John Zhuang Hui},
  title = {Formally Verified Memory Protection for a Commodity Multiprocessor Hypervisor},
  booktitle = {30th USENIX Security Symposium (USENIX Security 21)},
  year = {2021},
  isbn = {978-1-939133-24-3},
  pages = {3953--3970},
  url = {https://www.usenix.org/conference/usenixsecurity21/presentation/li-shih-wei},
  publisher = {USENIX Association},
  month = aug
}

@inproceedings{seL4Integrity,
  title = {{{seL4}} Enforces Integrity},
  booktitle = {Proceedings of the {{Second}} International Conference on {{Interactive}} Theorem Proving},
  author = {Sewell, Thomas and Winwood, Simon and Gammie, Peter and Murray, Toby and Andronick, June and Klein, Gerwin},
  year = 2011,
  month = aug,
  series = {{{ITP}} '11},
  pages = {325--340},
  publisher = {Springer-Verlag},
  address = {Berlin, Heidelberg},
  urldate = {2025-02-10},
  isbn = {978-3-642-22862-9}
}

@inproceedings{seL4InfoFlow,
  title = {{{seL4}}: From General Purpose to a Proof of Information Flow Enforcement},
  booktitle = {{{IEEE}} Symposium on Security and Privacy},
  author = {Murray, Toby and Matichuk, Daniel and Brassil, Matthew and Gammie, Peter and Bourke, Timothy and Seefried, Sean and Lewis, Corey and Gao, Xin and Klein, Gerwin},
  year = 2013,
  month = may,
  pages = {415--429},
  publisher = {IEEE},
  address = {San Francisco, CA},
  doi = {10.1109/SP.2013.35}
}

\end{document}